\documentclass[journal]{IEEEtran}
\usepackage{makecell}
\usepackage{supertabular}
\usepackage{multirow}
\usepackage{graphicx}
\usepackage{subfigure}
\usepackage[normalem]{ulem}
\usepackage{url}
\usepackage{bbm}
\usepackage{enumerate}
\usepackage{longtable}
\usepackage{booktabs}
\usepackage{multirow}
\usepackage{xspace}

\usepackage{multicol}
\usepackage{color}
\usepackage{threeparttable}
\usepackage{ulem}
\usepackage{comment}

\usepackage{float}
\usepackage{listings}
\usepackage{caption}
\usepackage{subcaption}
\usepackage{algorithm,algpseudocode}
\usepackage{amsmath}
\usepackage{amsfonts}
\usepackage{amssymb}

\newcommand{\eg}{\textit{e.g.,}\xspace}

\newcommand{\encoder}{HSTEncoder\xspace}

\newcommand{\Name}{$k$NN-MTS\xspace}
\newtheorem{definition}{Definition}
\usepackage{xcolor,soul,framed} %,caption

\colorlet{shadecolor}{yellow}
\usepackage{mathabx}
\usepackage{array}
\usepackage{mdwmath}
\usepackage{mdwtab}
\usepackage{eqparbox}
\usepackage{url}
\usepackage{fancyhdr} 
\begin{document}
\bstctlcite{IEEEexample:BSTcontrol}
    \title{Nearest Neighbor Multivariate Time Series Forecasting}
  \author{Huiliang Zhang,~\IEEEmembership{Student Member,~IEEE,}
Ping Nie,~\IEEEmembership{Student Member,~IEEE,} Lijun Sun, ~\IEEEmembership{Senior Member,~IEEE,} and~Benoit Boulet,~\IEEEmembership{Senior Member,~IEEE} 
      % <-this % stops a space

  \thanks{Huiliang Zhang and Benoit Boulet are with the Department of Electrical and Computer Engineering, McGill University, Montreal, QC H3A 0G4, Canada (e-mail: huiliang.zhang2@mail.mcgill.ca;
benoit.boulet@mcgill.ca; Lijun Sun is with the Department of Civil Engineering, McGill University, Montreal, QC H3A 0G4, Canada (e-mail: lijun.sun@mcgill.ca). Ping Nie is with the Peking University, CH 100871,
China (e-mail: ping.nie@pku.edu.cn).}
}

\markboth{IEEE  TRANSACTIONS ON NEURAL NETWORKS AND LEARNING SYSTEMS
% , VOL.~60, NO.~12, DECEMBER~2012
}{Zhang \MakeLowercase{\textit{et al.}}: Nearest Neighbor Multivariate Time Series Forecasting}

% ------- 版权行内容（自行替换年份、期刊、DOI） -------
\newcommand{\copyrightnotice}{%
  \scriptsize
  ©2024 IEEE. Personal use of this material is permitted.%
  \newline Published in \emph{IEEE Transactions on Neural Networks and Learning Systems}.%
  \newline DOI: 10.1109/TNNLS.2024.3490603%
}
% ---------- 首页页脚样式 ----------
\fancypagestyle{firstpage}{
  \fancyhf{}                              % 清空默认
  \renewcommand{\headrulewidth}{0pt}       % 去掉页眉横线
  \renewcommand{\footrulewidth}{0pt}       % 去掉页脚横线
  \fancyfoot[C]{\copyrightnotice}          % 页脚居中
}
% ====================================================================
\maketitle
\thispagestyle{firstpage}

% === ABSTRACT ====================================================================
% =================================================================================
\begin{abstract}
Multivariate Time Series (MTS) forecasting has a wide range of applications in both industry and academia. Recently, Spatial-Temporal Graph Neural Networks (STGNNs) have gained popularity as MTS forecasting methods. However, current STGNNs can only utilize the finite length of MTS input data due to the computational complexity. 
Moreover, they lack the ability to identify similar patterns throughout the entire dataset and struggle with data that exhibit sparsely and discontinuously distributed correlations among variables over an extensive historical period, resulting in only marginal improvements. In this paper, we introduce a simple yet effective $k$-nearest neighbor MTS forecasting (\Name) framework, which forecasts with a nearest neighbor retrieval mechanism over a large datastore of cached series, using representations from the MTS model for similarity search. This approach requires no additional training and scales to give the MTS model direct access to the whole dataset at test time, resulting in a highly expressive model that consistently improves performance, and has the ability to extract sparse distributed but similar patterns span over multi-variables from the entire dataset. 
Furthermore, a Hybrid Spatial-Temporal
Encoder (\encoder) is designed for \Name which can capture both long-term temporal and short-term spatial-temporal dependencies, and is shown to provide accurate representation for \Name for better forecasting.
% The proposed \encoder give the most informative representation from segment-level and point-level. 
Experiment results on several real-world datasets show a significant improvement in the forecasting performance of \Name. 
The quantitative analysis also illustrates the interpretability and efficiency of \Name, showing better application prospects and opening up a new path for efficiently using the large dataset in MTS models.

\end{abstract}

% === KEYWORDS ====================================================================
% =================================================================================

\IEEEpeerreviewmaketitle

% ====================================================================
% ====================================================================
% ====================================================================

% === I. INTRODUCTION =============================================================
% =================================================================================
\section{Introduction}
\IEEEPARstart{M}{ultivariate} time series (MTS) data is ubiquitous in our daily life, such as transportation~\cite{chen2023multi_scale,traffic_survey} and energy~\cite{energy_survey}.
Accurately forecasting future trends is incredibly valuable for 
decision-making,
% . Thus, MTS forecasting has been a persistent research topic for several decades
attracting attention from both academia and industry for several decades.
\begin{figure}[htbp!]
    \centering
    \setlength{\belowcaptionskip}{-0.4cm}
    \includegraphics[width=0.9\linewidth]{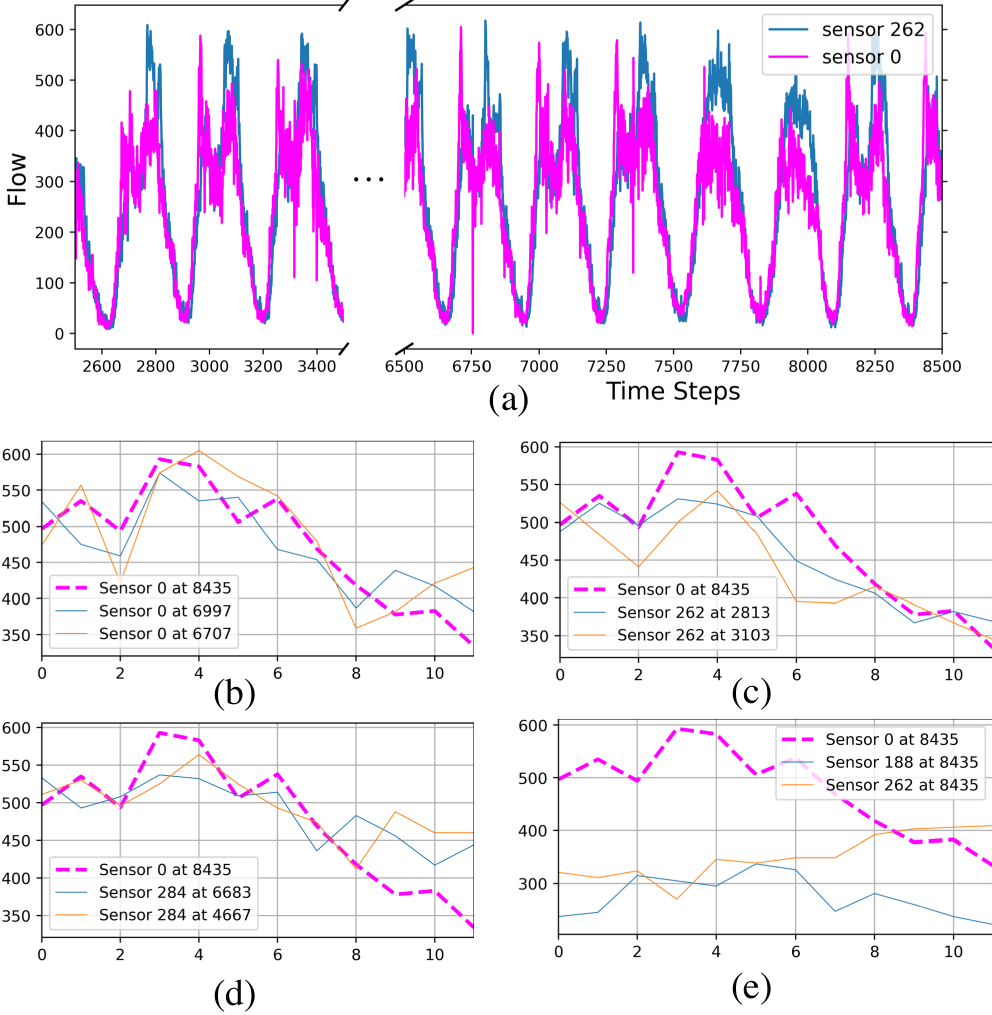}
    \caption{Traffic Flow Patterns from PEMS04 Dataset. (a) Adjacent sensors 0 and 262 showing complex spatial-temporal correlations. (b) Cyclical patterns of sensor 0. (c) Similar patterns from adjacent sensor 262 over time. (d) Similar patterns from non-adjacent sensor 284 over time. (e) Distinct patterns of sensor 0 and adjacent sensors 188 and 262 in a short window.
    } 
    \label{example}
\end{figure}

MTS data usually contains complex temporal and spatial relationships including similar patterns from different times and spatial dependencies from various nodes. Therefore, previous research typically formalizes MTS as spatial-temporal graph data~\cite{GWNet,2017DCRNN,stgcnyu2017spatio}.  Recently, Spatial-Temporal Graph Neural Networks (STGNNs) have gained popularity as MTS forecasting methods due to their cutting-edge performance \cite{stgcnyu2017spatio,MTS_survey}. 
In addition, emerging research in this field aims to construct more precise graph structure~\cite{stemgnncao2020spectral,shang2021discrete, ye2022learning}, fuse spatial-temporal data with elaborate graph neural network ~\cite{shao2022pre, GWNet,dgcrnli2021dynamic, stgcnyu2017spatio}, and utilize distinguishable spatial-temporal features~\cite{patchTst,stidshao2022spatial,JointSpatiotemporal_INNLS} or multi-scale features~\cite{Multiscale_INNLS, MR-Transformer_INNLS}. 

Despite diligent research in MTS forecasting, current methods are becoming more sophisticated but still rely on finite short-term or long-term input data 
% (\eg past 2 hours or 7 days) 
with graph structures. 
% This leads to limited improvements \cite{borgeaud2022improving,stidshao2022spatial}. 
Given the continuous growth in the volume of data 
% across spatial and temporal scales, including the increase 
in variables and length, existing MTS models with limited parameters face challenges in capturing the intricate spatial-temporal relationships present within the entire datasets, such as the cyclical patterns occurring from the super long history, and sparsely distributed cross-correlations among variables \cite{stidshao2022spatial,MTS_survey,TraverseNet_INNLS,he2024distributionalshift_INNLS}. 
%sparsely distributed discontinuous dependencies or cross-correlations among variables
% \hl{Need to further describe this current limitations, like self-periodicity, dependency from neighbor and others which have similar features in low dimension}
However, we argue that those cyclical patterns and correlations occurring from the whole datasets could have potential contributions to MTS forecasting.
% , considering the variable's similarity and reoccurring temporal patterns. 
We take the traffic flow system as an example to illustrate this argument.
% , where each sensor corresponds to a variable.  
Fig. \ref{example} (a) depicts traffic flows over two months from two adjacent sensors 0 (purple) and 262 (blue), with a sampling rate of 5 minutes per step. 
% The front and middle part of the figure has been omitted due to space limitations. 
Based on the very long-term historical data, we observe notable similarities and correlations between sensor 0 and sensor 262 within the ranges of 2.8k to 3.4k and 7k to 7.5k steps. Additionally, we notice similar patterns within sensor 0's history, particularly at the 6.7k and 6.9k steps, which resemble segments at the 8.4k step, and could be seen as cyclical patterns from the node itself, as shown in Fig. (b). 
Similarly, adjacent sensor 262 exhibits similar segments at 2.8k and 3.1k steps, as depicted in Fig. (c). These segments are sparsely and discontinuously distributed and span in the long history, illustrating the potential spatial dependencies and cross-correlations among variables.
Surprisingly, we find comparable segments from the non-adjacent sensor 284 at 6.6k and 4.6k steps, as illustrated in Fig. (d). It depicts the similar patterns that could occur from the sensors sharing the same characteristics in low dimensions and they are also cross-correlated. For instance, even when situated at different locations, two sensors at a central traffic intersection hub may exhibit analogous traffic dynamics at different times.
Counter-intuitively, sensor 0 displays significant differences in magnitude and trend compared to its adjacent sensors, 188 and 262, within the short time window at the 8.4k step, as shown in Fig. (e). This demonstrates the unreliability of current MTS methods relying solely on short-term history for forecasting. 
% \textcolor{blue}{The above observation shows the potential values of similar patterns occurring over the super-long history of the entire MTS dataset.}
Based on these observations, a natural idea for improving forecasting accuracy is to consider those similar and cyclical patterns occurring throughout the entire history and the cross-correlations
% underlying relations 
among variables. However, the valuable historical segments in the dataset are sparsely distributed in both temporal and spatial domains. Current methods, unfortunately, fall short in their capability to detect these patterns, even though these patterns could significantly boost forecasting outcomes.

Although massive valuable information and patterns are stored in the whole dataset, identifying and utilizing them for forecasting models present several challenges in MTS. First, the computational complexity increases linearly or even quadratically with the length of the input data~\cite{stgcnyu2017spatio,dgcrnli2021dynamic}. Furthermore, the computational burden intensifies with the number of variables when considering MTS simultaneously \cite{MTS_survey}. Second, the valuable historical information is sparsely distributed in the dataset, and shows complex dependencies across multi-variables, making it difficult to retrieve directly. To address these challenges, an early work \cite{knn1999k} on time series attempted to utilize $k$-Nearest Neighbors ($k$NN) for retrieving similar series based on raw input data to enhance forecasting. However, since it relies solely on univariate short-term raw time series data for retrieval without considering the spatial dependencies, it provides limited support for prediction. 
% This limitation stems from the inadequate amount of spatial and temporal information that is available.
Inspired by recent successful approaches in natural language processing (NLP), such as $k$NN-LM \cite{knnLM} and $k$NN-MT \cite{knnMT}, that retrieve encoded hidden states of the target token from training data to enhance NLP predictions, there may be potential benefits of retrieving encoded hidden states using $k$NN in MTS. 
However, forecasting MTS also encounters distinct challenges associated with spatial-temporal correlations, which renders the combination of $k$NN and MTS not an easy task. 
For example, MTS raw data are recorded separately by each node and do not explicitly contain spatial and temporal dependencies, so retrieving raw time series data can be difficult to reflect underlying spatial and temporal correlations. Furthermore, MTS forecasting is more complex due to the continuous space of possible values and multiple horizons (e.g., the next 12 time steps) in forecasting, which differs from NLP tasks where estimating the next word can be modeled as a multi-class classification problem. 
Therefore, further research is needed to explore successful retrieval and aggregation of informative historical time series in MTS forecasting.

In this paper, we introduce \Name, a non-parametric method for MTS forecasting that uses nearest neighbour retrieval to extract sparse but similar patterns from entire datasets via informative representations. 
Specifically, to address the computational challenges of utilizing the entire dataset,  \Name first transforms fixed-length historical MTS data into dense vectors in a datastore using the MTS model. For predictions, \Name retrieves similar representations from these vectors and fetches the corresponding future time series segments. The top 
$K$ similar segments are then aggregated for improved forecasting.
To capture the intricate spatial-temporal relationships from the entire dataset through representation retrieval, we further carefully design a Hybrid Spatial-Temporal
Encoder (\encoder) to incorporate both long-term temporal and short-term spatial-temporal dependencies together, which is shown to  benefit \Name most.
Furthermore, \Name can be added to any pre-trained MTS model without further training. We also find our \Name has understandable interpretability by inspecting the retrieved similar nodes, similar segments, and their time characteristics. For large time series data which could consume more computational resources to train an MTS model, \Name shows its training efficiency with comparable forecasting performance. 
In summary, the main contributions are the following:
\begin{itemize}
    \item We propose a simple and effective \Name framework, which is the first to utilize the nearest neighbor method through representations retrieval for MTS forecasting, demonstrating the potential of non-parametric methods with informative embeddings retrieval mechanism. \Name enables existing MTS models to directly access the entire dataset at test time without additional training, enhancing model expressiveness and consistently boosting performance by extracting sparse yet similarly distributed patterns across MTS.
    \item We carefully design a Hybrid Spatial-Temporal Encoder for \Name based on segmentation to capture both long-term temporal and short-term spatial-temporal dependencies. We explore different methods of MTS encoding and find all instantiations are helpful under the \Name framework. Retrieving representations from \encoder, which captures spatial-temporal correlations through long-term history, provides the best accuracy and overcomes computational challenges.
    \item We conduct extensive experiments on several real-world datasets to demonstrate the significant improvements in performance, interpretability, and efficiency of the proposed \Name method, showing how effective this approach can be for enhancing MTS forecasting.
\end{itemize}
The remaining part of this paper is organized as follows: Section \ref{sec:prelimi} presents the definitions used in \Name and formulates the research problem of MTS forecasting. Section \ref{sec:Methodology} illustrates the structures and key components of the \Name framework. Section \ref{sec:exp} and \ref{sec:abl} illustrate the research questions explored in this paper, and give details of the experimental setup and present the experimental results, analysis and ablation studies. Section \ref{sec:related} provides an overview of the related work, and section \ref{sec:conclusion} is the conclusion.

\section{Preliminaries}
\label{sec:prelimi}
This section presents the definitions used in \Name including multivariate time series, the segment-level representation, and the dependency graph. Then, we introduce the research problem of MTS forecasting.

\begin{definition} 
\textbf{Multivariate Time Series (MTS)}. 
MTS can be denoted as a tensor $\mathcal{X}\in\mathbb{R}^{T\times N\times C}$, where $T$ is the number of time steps, $N$ is the number of series, \eg the nodes or variables, and $C$ is the number of data channels for each series.
\end{definition}
\begin{definition}
\textbf{Segment-level Representation}. 
The segment-level representation can be denoted as a tensor $\mathcal{H}\in\mathbb{R}^{P\times N\times C}$, where $P$ is the number of segments. Each segment contains $L/P$ non-overlapping time steps, $L$ is the length of the input data.
\end{definition}
% The framework takes data with length $L$ as input and processes $N$ series simultaneously. 
\begin{definition}
\textbf{Dependency Graph}. 
In MTS forecasting, each variable depends not only on its own history but also on other variables. Such dependencies are captured by a dependency graph $\mathcal{G}=(V, E)$, where $V$ is the set of $|V|=N$ nodes, and each node corresponds to a variable. $E$ is the set of $|E|=M$ edges.
The adjacency matrix derived from a graph is denoted by $\mathbf{A} \in \mathbf{R}^{N\times N}$. 
The dependency graph could be predefined or learned through training.
\end{definition}

\begin{definition}
\textbf{MTS Forecasting}.
Given $N$
series with $T_h$ historical signals $\mathcal{X}\in\mathbb{R}^{T_h\times N\times C}$, MTS forecasting aims to predict next $T_f$ future values $\mathcal{Y} \in \mathbf{R}^{T_f \times N \times C}$, based on the dependency graph $\mathcal{G}$ and history signals $\mathcal{X}$.
\end{definition}

\begin{figure*}[htbp!]
 \centering
 \includegraphics[width=0.7\linewidth]{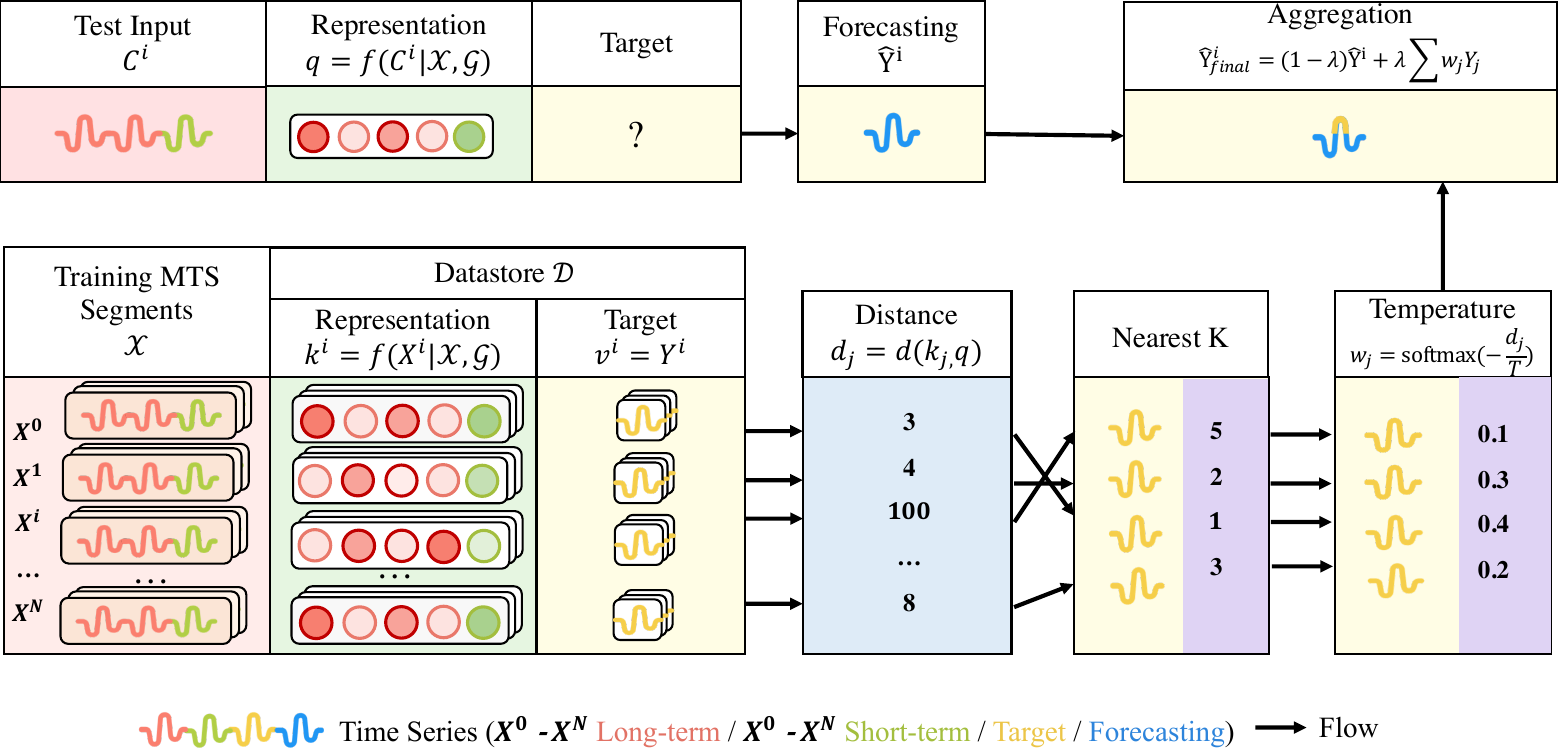}
 \caption{
 {\color{black}
 Architecture of \Name. The offline datastore $\mathcal{D}$ contains representations $f(X^i|\mathcal{X},\mathcal{G})$ of the training set $\mathcal{X}$ and target values $Y^i$. During forecasting, the query $q$ retrieves $K$ nearest keys from $\mathcal{D}$, which are weighted and combined with the model's output for final prediction.
}
}
\label{fig:framework}
\end{figure*}
% \section{NEAREST NEIGHBOR MULTIVARIATE TIME SERIES FORECASTING}
\section{Nearest Neighbor Multivariate Time Series Forecasting}
\label{sec:Methodology}
The \Name involves augmenting pre-trained MTS forecasting models, such as STGNNs, by incorporating a nearest neighbour retrieval mechanism 
% based on representations.
without requiring any additional training. 
The proposed \Name framework is shown in Fig. \ref{fig:framework}. 
% The proposed \Name framework is depicted in Figure \ref{fig:framework}. 
The history of the $i$-th series is represented as $X^{i} \in \mathbf{R}^{L \times C}$, and we use the $i$-th series as a forecasting target (represented by the blue curve) to illustrate the workflow of \Name.
\Name can be implemented with a single forward pass over a history data collection. The datastore is constructed offline from the original MTS training set $\mathcal{X}={\mathcal{X}^0, \mathcal{X}^1, ..., \mathcal{X}^N}$. The raw MTS data is transformed by a Hybrid Spatial-Temporal Encoder (\encoder) to obtain the MTS representations $f(X^i|\mathcal{X},\mathcal{G})$ as the key $k^i$. Subsequently, the representations and their corresponding target future values $Y^i$ pairs are stored in a key-value datastore. The representations remain unchanged during forecasting.
We hypothesize that time series that are closer in representation space are more likely to be followed by the same future series.
During forecasting, the test series $C^{i}$ from the $i$-th series is also transformed by \encoder and used as a query $q$ to retrieve the top $K$ nearest neighbours and their corresponding target future values $v$ from the datastore. The target values from the datastore are then weighted and summed with the original model's output forecasting.

% More specifically, given an input sequence from the MTS, a  xxxxx
% The \Name allows the model direct access to a datastore of cached examples.

% \vspace{-0.3cm}
\subsection{Datastore Creation}
\label{sec:data_store_create}
Our datastore is constructed offline and consists of a set of key-value pairs. The key is a high-dimensional representation 
% of the entire MTS, 
computed by the \encoder. Let $f(\cdot)$ be the function that maps a time series $X^i$ to a fixed-length vector representation. For example, $f(X^i|\mathcal{X},\mathcal{G})$ could map $X^i$ to an intermediate representation that is output from \encoder, considering multivariate input time series data and dependency graph.
Given the training examples $(X^i, Y^i) \in (\mathcal{X}^i,\mathcal{Y}^i)$ from series $i$, we define the key-value pair $(k^i, v^i)$, where the key is the vector representation $f(X^i|\mathcal{X},\mathcal{G})$, and the value is the corresponding target future series $Y^i$. For a parallel MTS collection $(\mathcal{X}, \mathcal{Y})$, the representations are generated by a single forward pass over each node. The complete datastore $(\mathcal{K}, \mathcal{V})$ is defined as follows:
\begin{align}
    \left( \mathcal{K}, \mathcal{V} \right) = \left \{(f(X^i|\mathcal{X},\mathcal{G}),~ Y^i), ~\forall{i} \in N ~|~ (X^i, Y^i) \in (\mathcal{X}, \mathcal{Y}) \right \}
\end{align}

\subsection{Forecasting}
At forecasting time, given the input series $C^i$, the model generates the output sequence $\hat{Y^i}$ and the representation $q = f(C^i|\mathcal{X},\mathcal{G})$. The model queries the data store with $q$ to retrieve its $K$-nearest neighbors $\mathcal{N}$, according to a distance function $d(\cdot,\cdot)$ (squared L2 distance in our experiments, making the similarity function an RBF kernel \cite{smola2004tutorial}). It then computes the weight over neighbours based on a softmax of their negative distances between the query and retrieved keys $k_j, j \in K$:

% while aggregating retrieved values for each series across all its occurrences in the retrieved targets (items that do not appear in the retrieved targets have zero weight):
% \begin{equation}
%      w_{k}(Y_k|C^i) \propto \sum_{(k_j,v_j) \in \mathcal{N}}
%      {\mathbbm{1}_{Y_i=v_j}\exp \left(\frac{-d(k_k,f(C^i|\mathcal{X})}{T}\right )}   
% \end{equation}
% \begin{equation}
%      w_{j}(Y_j|C^i) = \frac{\exp \left(\frac{-d(k_j,q)}{T}\right ) }{\sum\exp \left(\frac{-d(k_j,q)}{T}\right ) }
%      \label{equa:weight}
% \end{equation}
\begin{equation}
     w_{j}(Y_j|C^i) = \frac{\exp(z^j)}{\sum \exp(z^i)}, \quad z^j=\frac{-d(k_j,q)}{T}
     \label{equa:weight}
\end{equation}
% \begin{equation}
%     \begin{aligned}
%       w_{j}(Y_j|C^i) &= \frac{\exp(z^j)}{\sum \exp(z^i)} \\
%       z^j &=\frac{-d(k_j,q)}{T}   
%     \end{aligned}
%      \label{equa:weight}
% \end{equation}
where the $Y_j$ is the retrieved target value. A temperature $T$ greater than one flattens the weights and prevents overfitting to the most similar retrievals.
%Items not retrieved receive a probability of zero.
Although a pure $k$NN approach is effective, we improve the results by combining it with the base model prediction $\hat{Y}^i$, which is more robust in cases where there are no relevant cached examples. The model and $k$NN forecasting values are summed up to obtain the final \Name values:
\begin{align}
    \hat{Y}^i_\text{final} = (1-\lambda)~\hat{Y^i} + \lambda~\sum_{j\in K} w_{j}*Y_{j},
    \label{equa:aggre}
\end{align}
where  $\lambda = \frac{\alpha }{\bar{d}_j + \alpha }$ and $\bar{d}_j$ represents the average distance of the embeddings from $K$ retrieved representations to the current point. This enables the model to depend more on \encoder predictions when the retrieved neighbours are far from the target one. $\alpha$ is a constant that adjusts the scaling of $\lambda$.

\subsection{Representation Generation}
The high-dimensional representations are denoted as $f(\cdot)$ and utilized as keys in the retrieval process. The accuracy of the representations stored in these keys is critical to enable the query to efficiently locate the most relevant representations and corresponding target values. In addition, the informative nature of these representations facilitates a more precise direct prediction $\hat{Y}^i$. Therefore, it is crucial to design an encoder module which can generate a reasonable representation that effectively captures both periodicity and spatial information in the \Name.

Inspired by the recent Transformer-based works \cite{patchTst,zhang2022crossformer}, which show the segmentation of time series into subseries-level segments could retain local semantic information and enable the model to attend longer history, we design our Hybrid Spatial-Temporal Encoder (\encoder) for \Name to capture  long-term temporal dependency. Furthermore, the short-term spatial-temporal dependencies are also considered in \encoder for higher resolution information from the point-level in \Name's retrieval.

% \textbf{Univariate Temporal Representation}: Univariate temporal representation utilizes time series data from the node itself.
% , which means it only leverages the temporal information. 
Although long-term time series contain more temporal patterns, 
% the information density in the raw data is low, and 
it requires more computational resources to encode thousands of points directly in modern neural networks \cite{shao2022pre,patchTst}. Moreover, analyzing the patterns and dependencies of time series based on long-term historical MTS data is crucial since the occurrence of cyclical patterns is likely to be days, weeks even months. Therefore, we first slice the raw series data $X^{i}$ into non-overlapping sub-series and then transform them into temporal embeddings. Specifically, $X^{i}$ is divided into $P$ non-overlapping continuous segments $S = [S_0, S_1, ..., S_P]$, where each segment $S_j \in \mathbf{R}^{L_s \times C}$ contains $L_s$ time steps, and $L_s = L / P$. The temporal embedding of the $j$-th segment of $i$ series $\textbf{E}^i_j$ is generated by a linear layer with learnable positional embedding $\textbf{p}_j$:
\begin{equation}
    \mathbf{E}_j^i = \mathbf{W}\cdot S^i_j + \mathbf{b} + \mathbf{p}_j
\end{equation}
where $\mathbf{W}\in\mathbb{R}^{d\times L_s }$ and $\mathbf{b}\in\mathbb{R}^d$ are learnable parameters, $d$ is the hidden state dimension. 
% Then the temporal embeddings corresponding with positional embeddings are fed into 4 transformer layers to obtain context-aware segment-level representations.

%%%%%%%%%%%%%%%%%%%%%%%%%%%%%%%%%%%%%%%%%%%%%%%%%%%%%%%%%%%%%%%%%%%

As the segment embeddings only capture context within its data window of $L_s$, it can be challenging for the model to distinguish the significance of segments in various contexts. To address this issue, we incorporate multiple temporal Transformer layers to provide rich contextual information $\textbf{H}^i_j$ for the $j$-th segment of series $i$. Each temporal Transformer layer consists of a multi-head self-attention network (MSA) and a position-wise fully connected feed-forward network (FFN), with a residual connection and layer normalization:
\begin{equation}
 \begin{aligned}
 & \mathbf{U}^i=\mathrm{LayerNorm}(\mathbf{E}^i + \mathrm{MSA}(\mathbf{E}^i)) \\
& \mathbf{H}^i_{long}=\mathrm{LayerNorm}(\mathbf{U}^i + \mathrm{FFN}(\mathbf{U}^i))
 \end{aligned}
\label{eq:transformer}
\end{equation}
where $\mathbf{E}^i=(\mathbf{E}^i_1,\mathbf{E}^i_2,...\mathbf{E}^i_{P-1},\mathbf{E}^i_{P})$ are the input temporal embeddings. 
% Following STEP \cite{shao2022pre}, 
We use four stacked Transformer layers to get the final contextual representations which capture the long-term temporal dependencies from the segment level. 

In \Name, we also utilize the representations from the STGNNs to incorporate short-term spatial-temporal dependencies and enable the \encoder to have a higher resolution information from the point level. 
Since current STGNNs usually use the last historical segment before the predicted segment (\eg 12 timesteps for a segment) as input, the data we use to generate multivariate spatial-temporal representation could be the last segments $S_P$ from $X^i$.
Then we select the representation from a representative STGNN method Graph WaveNet~\cite{GWNet} to our \encoder, which combining graph convolution with dilated causal convolution could also enable the \Name to extract useful patterns considering the short-term spatial-temporal dependencies from the entire dataset.
% which combines graph convolution with dilated causal convolution.
% , Graph WaveNet can efficiently and effectively 

Mathematically, given a 1-D sequence input $S_p \in \mathbf{R}^{L_s \times C}$ and a filter $\mathbf{r} \in \mathbf{R}^{B\times C}$, the dilated causal convolution operation of $S_p$ with $\mathbf{r}$ at timestep $t$ is represented as
\begin{equation}
    \mathbf{H}=S\star \mathbf{l}(t) = \sum_{m=0}^{B-1} \mathbf{r}(m)S_{t-a\times m},
\end{equation}
where $a$ is the dilation factor which controls the skipping distance. 
% By stacking dilated causal convolution layers with dilation factors in an increasing order, the receptive field of a model grows exponentially. It enables dilated causal convolution networks to capture longer sequences with less layers, which saves computation resources. 
The hidden states $\mathbf{H}$ which encoded the short-term sequence information are then sent to the gated convolutional network to learn the temporal dependencies:
\begin{equation}
    \mathbf{Z} = g(\mathbf{\Theta_1}\star\mathbf{H}+\mathbf{c})\odot\sigma(\mathbf{\Theta_2}\star\mathbf{H}+\mathbf{d}),
\end{equation}
where $\mathbf{\Theta_1}$, $\mathbf{\Theta_2}$, $\mathbf{c}$ and $\mathbf{d}$ are model parameters, $\odot$ is the element-wise product, $g(\cdot)$ is an activation function of the outputs, and $\sigma(\cdot)$ is the sigmoid function which determines the ratio of information passed to the next layer. 
% We adopt Gated TCN in our model to learn complex temporal dependencies.  
% Although we empirically set the tangent hyperbolic function as the activation function $g(\cdot)$, other forms of Gated TCN can be easily fitted into our framework, such as an LSTM-like Gated TCN \cite{kalchbrenner2016neural}.
The following graph convolution layer is used to get the spatial dependencies:
\begin{equation}
    \textbf{H}_{short} = \sum_{k=0}^{B}\mathbf{P}_f^k\mathbf{Z}\mathbf{W}_{k1}+\mathbf{P}_b^k\mathbf{Z}\mathbf{W}_{k2}+\mathbf{\Tilde{A}}_{apt}^k\mathbf{Z}\mathbf{W}_{k3}.
\end{equation}
where $\mathbf{P}_f$ and $\mathbf{P}_b$ are the predefined forward and the backward adjacency matrix. If the predefined graph is missing, we could learn the adjacency matrix ${\Tilde{A}}_{apt}$ as dependency graph through training.
% when the graph structure is unavailable, the self-adaptive adjacency matrix $\mathbf{\Tilde{A}}$ is used alone to capture hidden spatial dependencies.

% \textbf{Hybrid Representation}: 
% In \Name, we utilize the representation from both univariate temporal representation and multivariate
% spatial-temporal representation to get a hybrid representation.  
For $i$-th time series, \Name leverages the short-term spatial-temporal hidden states $\textbf{H}^i_{short} \in \mathbf{R}^{1 \times d}$, the long-term temporal contextual representation $\textbf{H}^i_{long} \in \mathbf{R}^{1 \times d}$ from the transformer-based encoder to generate the fused hidden states $\textbf{H}^i_{hybrid} \in \mathbf{R}^{1 \times d}$:
\begin{equation}
\textbf{H}^i_{hybrid} = k^i = MLP(\textbf{H}^i_{short}) + MLP(\textbf{H}_{long}^i),
\label{equa:prediction1}
\end{equation}
then the fused hidden states $\textbf{H}^i_{hybrid}$ is sent to an MLP for forecasting:
\begin{equation}
    \hat{Y^i} = MLP(\textbf{H}^i_{hybrid}),
    \label{equa:prediction2}
\end{equation}
where the $\hat{Y^i} \in \mathbf{R}^{T_f}$ is the $T_f$ future points of $i$-th time series and $MLP$ is a Multi-Layer Perceptron. $\textbf{H}^i_{hybrid}$ provides both long-term contextual information and short-term spatial-temporal information to facilitate modelling dependencies between time series and could be used as keys and query in \Name.
% The hybrid representation generation method is named as \encoder, and $\textbf{H}^i_{hybrid}$ could be used as keys and query in \Name.
% The proposed \Name is a general framework which could enhance almost arbitrary MTS forecasting models. As we mentioned in section \ref{sec:data_store_create}, 

\begin{algorithm}
    \caption{\Name}
    \begin{algorithmic}[1]
        \Require{MTS $\mathcal{X}$, \text{retrieve nearest number}  $K$, \text{temperature} $T$, $\lambda$, \encoder $f(\cdot)$, initialize datastore as $\mathcal{D}=\emptyset$} 
        \Ensure{\Name output $\hat{Y}^i_{final}$ }
        \State Datastore creation:
            \For{$\mathcal{X}^i$ in $\mathcal{X}$}
                \For{$X^i$ in $\mathcal{X}^i$}
                \State Get $k^i = f(X^i|\mathcal{X})$ from \encoder and corresponding target future values $v^i$
                \State Store $k^i, v^i$ in $\mathcal{D}$
            \EndFor
            \EndFor
        \State Generate \textsc{Faiss} index using $\mathcal{D}$ 
        \State Forecasting:
            \For{$C^i$ in dataloader}
                \State Get query $ q= f(C^i|\mathcal{X})$ and prediction $ \hat{Y}^i = MLP(q)$ from \encoder         \Comment{Equ. (\ref{equa:prediction1}) and (\ref{equa:prediction2})}
                \State Retrieve top $K$ representations using \textsc{Faiss} index and get keys $v^i$ from  $\mathcal{D}$ using $q$              
                \State Calculate distance and weight  $w_0, w_1, ...,w_K$ for retrieved top $K$ segments                                   \Comment{Equ. (\ref{equa:weight})}
                \State Calculate $\hat{Y}^i$ and $\hat{Y}^i_{final}$  \Comment{Equ. (\ref{equa:aggre})}
            \EndFor
    \end{algorithmic}
    \label{algo:knn_MTS}    
\end{algorithm}

\subsection{Space and Computation Cost}
Algorithm \ref{algo:knn_MTS} outlines the pseudo-code of \Name. The offline datastore is first constructed. During forecasting, the query representation from the test input encoded by the \encoder is used to retrieve the $K$ nearest representations (keys) from the datastore, and the retrieved target values are then weighted and summed with the original model's output forecast to obtain the final prediction.
Although storage space is required for the datastore in the \Name, it is not GPU-based, making this storage cost-effective. The reason is that the \Name does not introduce any trainable parameters and does not necessitate additional parameter updates on the GPU. The main computation cost of constructing the datastore is a single forward pass over all examples, which is only a fraction of the cost of training on the same examples for one epoch. Additionally, querying datastore could also be performed using \textsc{Faiss} \cite{faiss} in practice, a library that allows for fast nearest neighbour search 
in high-dimensional spaces to improve generation speed. More details about space and computational cost are shown in Sec. \ref{RQ7}. 

\section{Experiments}
\label{sec:exp}
In this section, we first cover the experimental settings. 
% of the datasets, baselines, evaluation metrics, and implementations. 
Then we present the experimental results of our proposed \Name on several real-world datasets. Moreover, we investigate the following research questions (RQ), with the experiments and analysis showing in the main results, visualizations and ablation study:
%
% the impact of each module in \Name. We also provide a detailed comparison between different model settings to demonstrate the forecasting performance of \Name.
%
% More details such as dataset and baselines' descriptions, optimization settings, and more visualizations and efficiency study can be found in the Appendix. 

\begin{itemize}
    \item RQ1: How does \Name perform on real-world MTS datasets compared with the state-of-the-art baselines?
    \item RQ2: What do the intermediate results of \Name look like including  retrieved representations, target values (segments) and corresponding sensors? Can these results further enhance the interpretability of \Name?
    \item RQ3: What do the predicted time series of \Name look like compared with the state-of-the-art baseline?
    \item RQ4: What are the impacts of employing different representations (encoders) in \Name? Does it show the generality of \Name?
    % merely consider short-term spatial-temporal correlations or long-term temporal dependency?
    \item RQ5: Does \Name improve the training efficiency of spatial-temporal models and show its expressiveness?
    \item RQ6: What are the effects of different hyper-parameters in \Name?
    \item RQ7: What's the computation and space cost of \Name?
    
\end{itemize}

\subsection{Experimental Setups }
\label{exp_set}
\subsubsection{Datasets}
\begin{table*}[ht!]
% \setlength{\abovecaptionskip}{-0.05cm}
% \setlength{\belowcaptionskip}{-0.3cm}
% \setstretch{1.18}
\caption{Statistics of datasets.}
\centering
\label{tab:datasets}
\scalebox{1}{
  \begin{tabular}{c|c|c|c|c|c}    % Dataset Samples Nodes SampleRate TimeSpan
    \toprule
    \textbf{Dataset}&\textbf{Type} &\textbf{\# Samples} & \textbf{\# Node} & \textbf{Sample Rate} & \textbf{Time Span} \\
    % \midrule
    \midrule
    % {METR-LA}  & 34272 & 207 &5mins & 4 months\\
    {PEMS-BAY} & Traffic speed & 52116 & 325 &5mins & 6 months\\
    {PEMS04}   & Traffic flow & 16992 & 307 &5mins & 2 months\\
    {PEMS07}   & Traffic flow & 28224 & 883 &5mins & 3 months\\
    {PEMS08}   & Traffic flow & 17856 & 170 &5mins & 2 months\\
    {Electricity} & Electricity consumption & 2208 & 336 & 60mins & 3 months\\
    \bottomrule
  \end{tabular}
  }
  % \vspace{-0.3cm}
\end{table*}
Following previous works~\cite{ shao2022pre, 2021GTS, 2020MTGNN, GWNet, stidshao2022spatial}, we conduct experiments on five commonly used multivariate time series datasets: PEMS-BAY, PEMS04, PEMS07, PEMS08 and Electricity, and most of them have tens of thousands of time steps and hundreds of sensors. 
% PEMS-BAY is a traffic speed dataset collected from California Transportation Agencies (CalTrans) Performance Measurement System (PeMS) \cite{PEMS-BAY}.
PEMS-BAY is the traffic speed datasets, while PEMS04, 07, 08 are traffic flow dataset. Traffic speed data records the average vehicle speed (miles per hour). Due to the speed limit in these areas, the speed data is a float value usually less than 70. The flow data should be an integer, up to hundreds, because it records the number of passing vehicles. All these datasets have one feature channel, i.e. $C=1$,
% PEMS-BAY is a publicly available traffic speed dataset collected from the California Transportation Agencies (CalTrans) Performance Measurement System (PeMS) \cite{PEMS-BAY}. It contains data from 325 sensors in the Bay Area over a 6-month period, from January 1st, 2017 to May 31st, 2017, as used in \cite{2017DCRNN}. PEMS04, PEMS07 and PEMS08 are traffic flow dataset collected from CalTrans PeMS (January 1st, 2018 to February 28th, 2018), Bay Area (May 1st, 2017 to August 31st, 2018) and District08 (July 1st, 2018 to August 31st, 2018). 
The statistics of five datasets are summarized in Table \ref{tab:datasets}. 
For a fair comparison, we follow the dataset division in previous works \cite{GWNet,stidshao2022spatial}. The ratio of training, validation, and test sets for the PEMS-BAY dataset is 7:1:2, while the ratio for other datasets is 6:2:2. It's notable that PEMS-BAY, PEMS04, PEMS07 and PEMS08 come with a pre-defined graph and Electricity is not.

\begin{comment}
\begin{itemize}
    \item \textbf{PEMS-BAY}, which is a publicly available traffic speed dataset collected from the California Transportation Agencies (CalTrans) Performance Measurement System (PeMS) \cite{PEMS-BAY}. It contains data from 325 sensors in the Bay Area over a 6-month period, from January 1st, 2017 to May 31st, 2017, as used in \cite{2017DCRNN}.
    \item \textbf{PEMS04}, which is a traffic flow dataset collected from CalTrans PeMS and contains data from 307 sensors in the Bay Area over 2 months, from January 1st, 2018 to February 28th, 2018, as used in \cite{2019ASTGCN}.
    \item \textbf{PEMS07}, which is a traffic flow dataset containing data from 883 sensors in the Bay Area over a 3-month period, from May 1st, 2017 to August 31st, 2018, as used in \cite{2019ASTGCN}.
    \item \textbf{PEMS08}, which is a traffic flow dataset containing data from 170 sensors in District08 over a 2-month period, from July 1st, 2018 to August 31st, 2018, as used in \cite{2019ASTGCN}.
\end{itemize}
\end{comment}
\subsubsection{Baselines}
We select various baselines from traditional to deep learning based models that have official public codes. Historical Average (HA), VAR ~\cite{VAR}
% , and SVR~\cite{SVR} 
are traditional methods. FC-LSTM~\cite{2014Seq2Seq}, DCRNN~\cite{2017DCRNN}, Graph WaveNet~\cite{GWNet}, 
% ASTGCN~\cite{2019ASTGCN}, 
% STSGCN~\cite{2020STSGCN}, 
and STGCN~\cite{stgcnyu2017spatio} are deep learning methods.
 GMAN~\cite{2020GMAN}, MTGNN~\cite{2020MTGNN}, MRN-CSG~\cite{Multiscale_INNLS}, STEP~\cite{shao2022pre}, PatchTST~\cite{patchTst}, STID~\cite{stidshao2022spatial} are recent state-of-the-art
works.

\subsubsection{Metric}
All baselines are evaluated by three
commonly used metrics in multivariate time series forecasting,
including Mean Absolute Error (MAE), Root Mean Squared Error
(RMSE) and Mean Absolute Percentage Error (MAPE). 
% Their formulas are as follows:
% \begin{equation}
% \begin{aligned}
%     & \text{MAE}(x, \hat{x})=\frac{1}{|\Omega|} \sum_{i\in\Omega}|x_i - \hat{x}_i|, \\
%     & \text{RMSE}(x, \hat{x})=\sqrt{\frac{1}{|\Omega|} \sum_{i\in\Omega}(x_i - \hat{x}_i)^2}, \\
%     & \text{MAPE}(x, \hat{x})=\frac{1}{|\Omega|} \sum_{i\in\Omega}\frac{|x_i - \hat{x}_i|}{x_i},
% \end{aligned}
% \end{equation}
% where $x_i$ denotes the $i$-th ground truth, while $\hat{x}_i$ represents the $i$-th predicted value. Furthermore, $\Omega$ is the set of indices of observed samples and $|\Omega|=T_f=12$. Mean Absolute Error (MAE) metric reflects the overall prediction accuracy, as mentioned in the reference~\cite{2021DGCRN}, the Root Mean Square Error (RMSE) metric is more sensitive to abnormal values in the data. On the other hand, the Mean Absolute Percentage Error (MAPE) can eliminate the influence of data units to some extent, making it a suitable choice for comparing the performance of models on different datasets. 

\subsubsection{Implementation} We conducted experiments to forecast the next 12 time steps using the same settings as the baselines. 
\encoder uses an input length of $L=2016$ in PEMSBAY, 04, 07, 08 datasets and  $L=288$ in  Electricity, which corresponds to the data from the past week. We set the input length for STGNN to $L_s=12$.
% and the number of segments to $P=168$ for all datasets. 
We used 4 layers of Transformer blocks in \encoder with a latent representation dimension of $d=96$. For Graph WaveNet, we used the hyper-parameters described in the original paper \cite{GWNet}, when the graph structure is unavailable, the self-adaptive adjacency matrix $\mathbf{\Tilde{A}}$ is used alone to capture hidden spatial dependencies.
To perform the forecasting, we retrieved 50 nearest neighbour representations and their target values and weighted their sum with the original model's prediction. We set $T$ to 1 and $\alpha$ to 0.2. 
% We used the default setting in STEP \cite{shao2022pre} to obtain the pre-trained \encoder, 
For \encoder, we set Adam as the optimizer and learning rates of 0.001, 0.001, 0.002, 0.001, and 0.001 for PEMS-BAY, PEMS04, PEMS07, PEMS08, Electricity respectively. The batch size was set to 16. We initialized the positional embeddings with a uniform distribution and used a truncated normal distribution with $\mu=0$ and $\sigma=0.02$ to initialize the last token, following the approach used in MAE \cite{2021MAE}. 
We also implemented the Transformer blocks using the PyTorch official implementation and performed significance tests (t-test with a p-value $<$ 0.05) on the experimental results. 
We use \textsc{Faiss} to search over the datastore, and found in preliminary experiments that using squared L2 distance for \textsc{Faiss} retrieval results in better performance for \Name, compared to inner product distance.
All experiments were run on a server equipped with RTX 6000 graphics cards and 24 GB of RAM. 
Our code can be found in this repository\footnote{https://github.com/hlhang9527/KNN-MTS}.

\subsection{Main Results (RQ1)}
\begin{table*}[htpb]
\renewcommand\arraystretch{1}
    \centering
    \caption{Multivariate time series forecasting on the PEMS-BAY,  PEMS04, PEMS07, PEMS08 and Electricity datasets. Numbers marked with $^*$ indicate that the improvement is statistically significant compared with the best baseline ~(t-test with p-value$<0.05$)}.
    \begin{tabular}{ccccr|ccr|ccr}
      \toprule
      \midrule
      \multirow{2}*{\textbf{Datasets}} &\multirow{2}*{\textbf{Methods}} & \multicolumn{3}{c}{\textbf{Horizon 3}} & \multicolumn{3}{c}{\textbf{Horizon 6}}& \multicolumn{3}{c}{\textbf{Horizon 12}}\\ 
      \cmidrule(r){3-5} \cmidrule(r){6-8} \cmidrule(r){9-11}
      &  & MAE & RMSE & MAPE & MAE & RMSE & MAPE & MAE & RMSE & MAPE\\
    \midrule
    \multirow{14}*{\textbf{PEMS-BAY}} 
      &HA              & 1.89  & 4.30  & 4.16\%        & 2.50  & 5.82  & 5.62\%       & 3.31  & 7.54  & 7.65\% \\ 
      &VAR             & 1.74  & 3.16  & 3.60\%        & 2.32  & 4.25  & 5.00\%       & 2.93  & 5.44  & 6.50\% \\  
      &FC-LSTM         & 2.05  & 4.19  & 4.80\%        & 2.20  & 4.55  & 5.20\%       & 2.37  & 4.96  & 5.70\% \\ 
      &DCRNN           & 1.38  & 2.95  & 2.90\%        & 1.74  & 3.97  & 3.90\%       & 2.07  & 4.74  & 4.90\% \\ 
      &STGCN           & 1.36  & 2.96  & 2.90\%        & 1.81  & 4.27  & 4.17\%       & 2.49  & 5.69  & 5.79\% \\ 
      &Graph WaveNet   & 1.30  & 2.74  & 2.73\%        & 1.63  & 3.70  & 3.67\%       & 1.95  & 4.52  & 4.63\% \\
      &GMAN            & 1.34  & 2.91  & 2.86\%        & 1.63  & 3.76  & 3.68\%       & 1.86  & 4.32  & 4.37\% \\  
      &MTGNN           & 1.32  & 2.79  & 2.77\%        & 1.65  & 3.74  & 3.69\%       & 1.94  & 4.49  & 4.53\% \\  
       &MRN-CSG         & 1.30  & 2.74  & 2.73\%       & 1.72  & 3.89  & 3.87\%       & 2.11  & 4.74  & 4.86\% \\    
      &STID            & 1.30  & 2.81  &2.73\%         & 1.62  & 3.72  & 3.68\%       & 1.89  & 4.40  & 4.47\% \\
      &STEP            & 1.26  & 2.73  &2.61\%        & 1.55  &3.58  & 3.45\%       & 1.79  & 4.20  & 4.18\% \\     
      &PatchTST          & 1.28  & 2.77  &2.71\%         & 1.60  & 3.63  & 3.61\%       & 1.89  & 4.42  & 4.43\% \\      
    \cmidrule(r){2-11}
            &\Name & \textbf{1.23}$^*$  & \textbf{2.64}$^*$  & \textbf{2.58\%}$^*$        & \textbf{1.52}  & \textbf{3.49}$^*$  & \textbf{3.38\%}$^*$      & \textbf{1.74}  & \textbf{4.15}$^*$  & \textbf{4.07\%}$^*$ 
      \\ 
    \midrule  
    % \midrule
    \color{black}{\multirow{14}*{\textbf{PEMS04}}}
      &HA              & 28.92  & 42.69  & 20.31\%        & 33.73  & 49.37  & 24.01\%       & 46.97  & 67.43  & 35.11\% \\ 
      &VAR             & 21.94  & 34.30  & 16.42\%        & 23.72  & 36.58  & 18.02\%        & 26.76  & 40.28  & 20.94\% \\ 
      % &SVR             & 22.52  & 35.30  & 14.71\%        & 27.63  & 42.23  & 18.29\%       & 37.86  & 56.01  & 26.72\% \\ 
      &FC-LSTM         & 21.42  & 33.37  & 15.32\%        & 25.83  & 39.10  & 20.35\%       & 36.41  & 50.73  & 29.92\% \\ 
      &DCRNN           & 20.34  & 31.94  & 13.65\%        & 23.21  & 36.15  & 15.70\%       & 29.24  & 44.81  & 20.09\% \\ 
      &STGCN           & 19.35  & 30.76  & 12.81\%        & 21.85  & 34.43  & 14.13\%       & 26.97  & 41.11  & 16.84\% \\ 
      &Graph WaveNet   & 18.15  & 29.24  & 13.54\%        & 19.12  & 30.58  & 14.27\%       & 20.69  & 33.02  & 15.47\% \\
      &GMAN            & 18.28  & 29.32  & 12.35\%        & 18.75  & 30.77  & 12.96\%       & 19.95     & 31.21  & 12.97\% \\  
      &MTGNN           & 18.22  & 30.13  & 12.47\%        & 19.27  & 32.21  & 13.09\%       & 20.93  & 34.49  & 14.02\% \\  
       &MRN-CSG        & 19.10  & 30.72  & 13.55\%        & 20.07  & 33.41  & 15.76\%       & 25.71  & 37.64  & 16.79\% \\   
      &STID           & 17.51  & 28.48   & 12.00\%        & 18.29  & 29.86  & 12.46\%       & 19.58  & 31.79  & 13.38\% \\
      &STEP           & 17.34  & 28.46   &12.10\%         & 18.24 & 29.81    &12.49\%       & 19.34 & 31.52   & 13.26\% \\ 
      &PatchTST        & 17.42  & 28.67   & 11.96\%        &  18.23 & 29.91  & 12.34\%      & 19.41  & 31.71  & 13.09\% \\      
    \cmidrule(r){2-11}
     & \Name & \textbf{17.08}$^*$  & \textbf{27.90}$^*$ & \textbf{11.52}\%$^*$       & \textbf{17.75}$^*$  & \textbf{29.32}$^*$ & \textbf{11.89}\%$^*$   & \textbf{18.68}  & \textbf{30.69} & \textbf{12.59}\%$^*$ \\
        % \midrule    
    \hline
    \color{black}{\multirow{14}*{\textbf{PEMS07}}}
      &HA              & 49.02   & 71.16   & 22.73\%     & 49.03   & 71.18   & 22.75\%   & 49.06   & 71.20   & 22.79\% \\ 
      &VAR             & 32.02   & 48.83   & 18.30\%     & 35.18   & 52.91   & 20.54\%   & 38.37   & 56.82   & 22.04\% \\ 
      % &SVR             & 30.15   & 42.41   & 19.22\%     & 37.61   & 49.00   & 19.35\%   & 37.76   & 51.90   & 23.19\% \\ 
      &FC-LSTM         & 20.42   & 33.21   & 8.79\%       & 23.18   & 37.54   & 9.80\%    & 28.73   & 45.63   & 12.23\% \\ 
      &DCRNN           & 19.45   & 31.39   & 8.29\%      & 21.18   & 34.43   & 9.01\%     & 24.14   & 38.84   & 10.42\% \\ 
      &STGCN           & 20.33   & 32.73   & 8.68\%      & 21.66   & 35.35   & 9.16\%    & 24.16   & 39.48   & 10.26\% \\ 
      &Graph WaveNet   & 18.69   & 30.69   & 8.45\%       & 20.24   & 33.32   & 8.57\%     & 22.79   & 37.11   & 9.73\% \\
      &GMAN            & 19.25   & 31.20   & 8.21\%       & 20.33   & 33.30   & 8.63\%     & 22.25   & 36.40   & 9.48\%    \\
      
      &MTGNN           & 19.23  & 31.15  &  8.55\%        & 20.83  & 33.93  &  9.30\%       & 23.60  & 38.13  & 10.17\% \\ 
       &MRN-CSG          & 19.79    & 31.20   & 8.11\%      & 20.97   & 34.02   & 9.00\%     & 23.97   & 37.53   & 11.05\% \\ 
      &STID            & 18.71  & 30.59  & 8.42\%         & 19.59  & 32.90  &  8.50\%       & 21.52  & 36.29  &  9.50\%      \\
      &STEP            & 18.56   & 30.33   & 8.35\%       & 19.69   & 32.55   & 8.56\%     & 21.79   & 35.30   & 9.86\% \\
      &PatchTST          & 18.60  & 30.42  &  8.79\%        & 20.23  & 33.21  &  8.53\%       & 22.30  & 36.68  & 9.59\%        \\      
    \cmidrule(r){2-11}
            &\Name          & \textbf{18.18}$^*$  & \textbf{30.19}$^*$  & \textbf{7.75}\%$^*$  & \textbf{19.32}$^*$  & \textbf{32.27}$^*$  & 
      \textbf{8.40}\%    & \textbf{20.34}$^*$  & \textbf{34.41}$^*$  & \textbf{9.32\%}$^*$ \\ 
        % \hline   
        % \midrule  
    \hline
    \color{black}{\multirow{14}*{\textbf{PEMS08}}}
      &HA              & 23.52  & 34.96  & 14.72\%        & 27.67  & 40.89  & 17.37\%       & 39.28  & 56.74  & 25.17\% \\ 
      &VAR             & 19.52  & 29.73  & 12.54\%        & 22.25  & 33.30  & 14.23\%        & 26.17  & 38.97  & 17.32\% \\ 
      % &SVR             & 17.93  & 27.69  & 10.95\%        & 22.41  & 34.53  & 13.97\%       & 32.11  & 47.03  & 20.99\% \\ 
      &FC-LSTM         & 17.38  & 26.27  & 12.63\%        & 21.22  & 31.97  & 17.32\%       & 30.69  & 43.96  & 25.72\% \\ 
      &DCRNN           & 15.64  & 25.48  & 10.04\%        & 17.88  & 27.63  & 11.38\%       & 22.51  & 34.21  & 14.17\% \\ 
      &STGCN           & 15.30  & 25.03  &  9.88\%        & 17.69  & 27.27  & 11.03\%       & 25.46  & 33.71  & 13.34\% \\ 
      &Graph WaveNet   & 14.02  & 22.76  &  8.95\%        & 15.24  & 24.22  &  9.57\%       & 16.67  & 26.77  & 10.86\% \\
    &GMAN            & 13.80  & 22.88  &  9.41\%        & 14.62  & 24.02  & 9.57\%       & 15.72  & 25.96  & 10.56\% \\
      &MTGNN           & 14.24  & 22.43  &  9.02\%        & 15.30  & 24.32  & 9.58\%       & 16.85  & 26.93  & 10.57\% \\  
    &MRN-CSG         & 14.39  & 24.63  & 10.01\%        & 17.52  & 26.89  & 11.05\%       & 21.56  & 30.13  & 14.21\% \\   
      &STID            & 13.28  & 21.66  &  8.62\%        & 14.21  & 24.57  & 9.35\%       & 15.58  & 25.89  & 10.33\%   \\
      &STEP            & 13.24   & 21.37  & 8.71\%         & 14.04  & 24.03 & 9.46\%       & 15.01  & 24.89   & 9.90\%\\
      &PatchTST          & 13.31  & 21.65  &  8.65\%        & 14.09  & 23.93  & 9.41\%       & 15.53  & 26.36  & 10.05\%   \\      
    \cmidrule(r){2-11}
      &\Name           & \textbf{12.64}$^*$  & \textbf{20.93}$^*$  & \textbf{8.45\%}         & \textbf{13.30}$^*$  & \textbf{22.39}$^*$  & \textbf{9.12\%}$^*$      & \textbf{14.11}$^*$  & \textbf{23.82}$^*$  & \textbf{9.69\%}$^*$ \\       
        % \hline   
        \midrule  
    \color{black}{\multirow{14}*{\textbf{Electricity}}}
      &HA              & 92.44  & 167.00  & 70.16\%        & 92.85  & 167.05  & 70.46\%       & 92.79  & 167.21  & 70.91\% \\ 
      &VAR             & 27.69  & 56.06  & 75.53\%        & 28.19  & 57.55  & 79.94\%        & 29.34  & 60.45  & 86.62\% \\ 
      &FC-LSTM         & 18.57  & 48.86  & 32.88\%        & 20.68  & 48.96  & 37.21\%       & 23.79  & 56.44  & 39.98\% \\ 
      &Graph WaveNet             & 21.45  & 41.09  &  57.12\%        & 23.56  & 46.95  & 63.34\%       & 24.98  & 51.97  & 62.81\% \\
      &MTGNN           & 16.78  & 36.91  &  48.16\%        & 18.43  & 42.62  & 51.31\%       & 20.49  & 48.33  & 56.25\% \\  
    &MRN-CSG          & 18.92  & 42.13  & 32.62\%        & 19.08  & 46.19  & 37.99\%       & 22.21  & 50.28  & 40.06\% \\ 
      &STID            & 16.21  & 35.23  &  32.87\%        & 18.31  & 42.12  & 37.80\%       & 19.59  & 47.41  & 43.02\%   \\
      &STEP            & 16.07  & 34.49  &  31.95\%        & 17.87  & 41.65  & 37.83\%       & 19.25  & 45.77  & 40.26\%   \\      
      &PatchTST         & 17.13  & 35.47  &  32.12\%        &18.36  &42.18   &37.98\%  & 19.69  & 45.89       & 41.30\%   \\      
    \cmidrule(r){2-11}     
        &\Name           & \textbf{15.85}$^*$  & \textbf{34.19}  & \textbf{31.40\%}      & \textbf{17.25}  & 
        \textbf{40.76}  & 
        \textbf{37.78\%}      & \textbf{19.03}$^*$  & \textbf{44.28}$^*$  & \textbf{39.45}\%$^*$ \\   
      \bottomrule
    \end{tabular}
    \label{tab:main}
    % \vspace{-4mm}
  \end{table*}
We compare the performance of \Name and all baseline methods on the 3rd, 6th and 12th time slots, respectively. The best results are highlighted in bold. In addition, DCRNN, STGCN and GMAN rely on a pre-defined graph so the results of these methods are not available in the Electricity dataset. 
Table \ref{tab:main} presents the main results of all baselines and \Name,  and 
\Name consistently outperforms the other models across most horizons in all datasets, confirming its effectiveness. We attribute this to the \Name framework's ability to extract and aggregate sparse distributed but similar patterns span over multi-variables from the entire dataset with $k$-nearest neighbors.
% aggregate similar segments , 
Specifically, traditional methods such as HA and VAR perform poorly due to their rigid assumptions about the data, such as stationarity or linearity. FC-LSTM, a classic recurrent neural network for sequential data, and MRN-CSG, a multi-scales RNN-based method also fall short as they only consider temporal features and ignore the spatial dependencies between MTS. Recently proposed spatial-temporal models, including STGCN, DCRNN, and Graph WaveNet, GMAN, MTGNN have overcome these shortcomings and made significant progress, but are still limited to the short-term point-level input data. In the STID method, the spatial and temporal identity information is captured using MLP, but it fails to track the exact message flow inside MTS. STEP and PatchTST utilize Transformer-based method to enable the model to detect the long-term temporal dependencies, achieving better performance than other baselines. However, the finite input prevents them from extracting all potential input-output relationships which are sparsely distributed over multi-variables from the entire dataset.  In conclusion, the results in Table~\ref{tab:main} validate the superiority of \Name, offering stable performance gains for current MTS forecasting models by having direct access to the whole dataset.
% sparse distributed but similar patterns span over multi-variables from the entire dataset, especially those  cyclical patterns from the node itself, . 
\begin{figure}[h!]
    \centering
    \setlength{\abovecaptionskip}{-0.0cm}
    \includegraphics[width=0.8\linewidth]{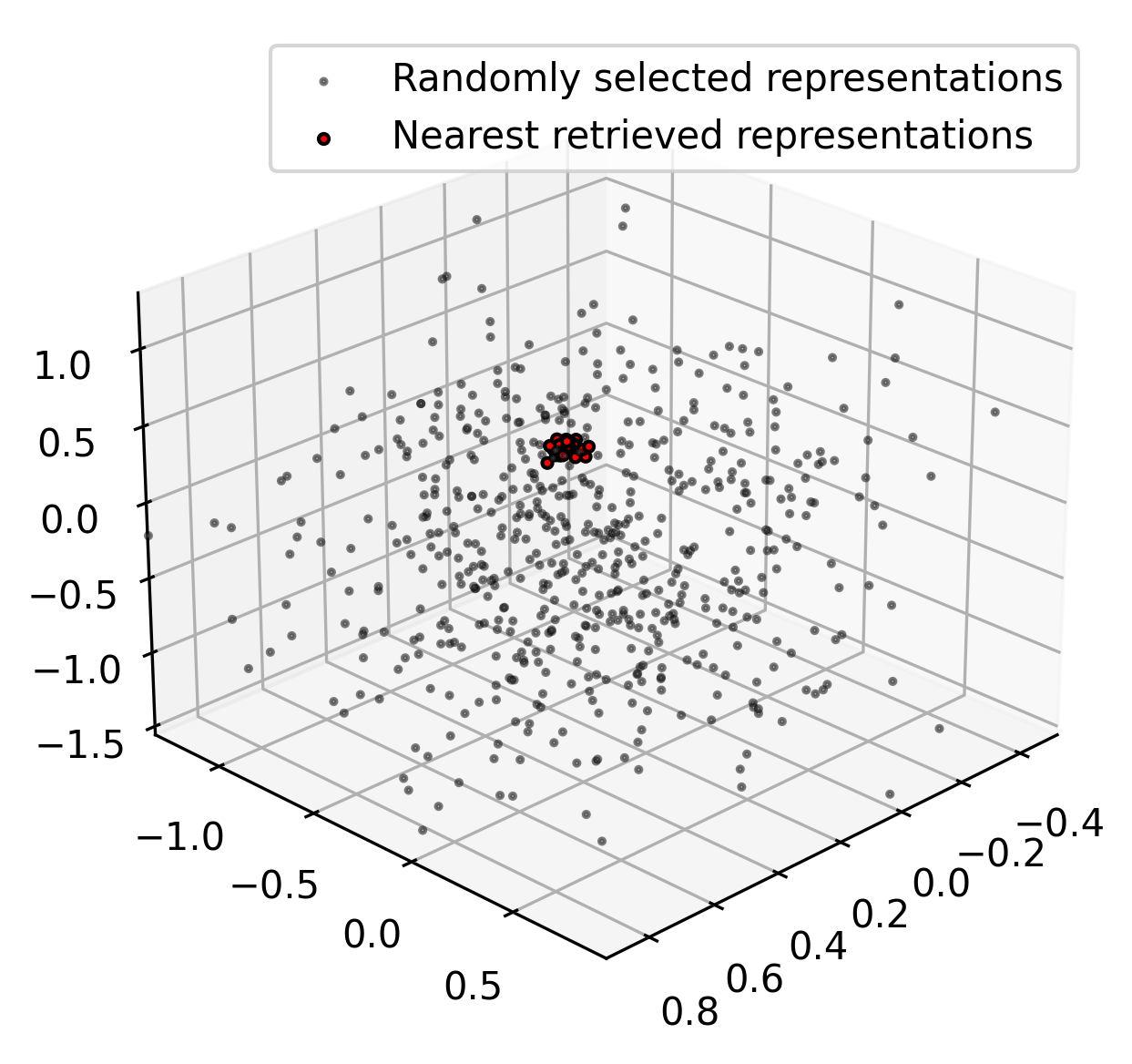}
    \caption{Visualization of retrieved nearest representations in lower three-dimensional space.  The retrieved representations cluster closely together
    while randomly sampled representations from the datastore are distant from each other.}
    \label{fig:retrival_embeding}
\end{figure}

\subsection{Visualization (RQ2 and RQ3)}
To intuitively understand the useful information retrieved by \Name from the whole dataset, in this subsection, we inspect the retrieved representations, similar segments and corresponding sensors from \Name,
and also show the significant contribution of these retrieved segments to the final prediction. We conduct experiments on the PEMS04 test dataset and randomly select a sample from it. 
\begin{figure*}[htbp]
 \centering
 \includegraphics[width=1\linewidth]{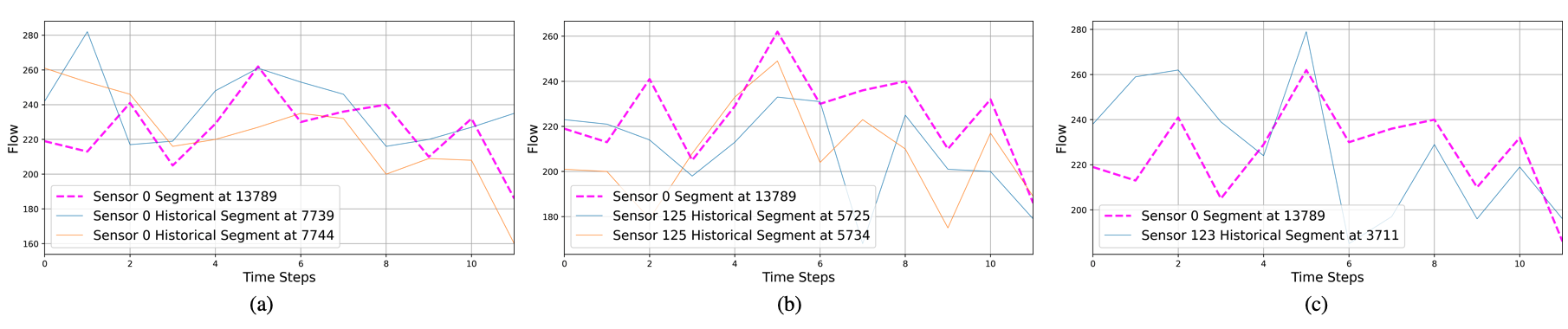}
 \caption{Retrieved similar historical segments. (a) Similar segments from the same sensor 0. (b) Similar segments from adjacent sensor 125. (c) Similar segments from non-adjacent sensor 123. }
 \label{vis_retr_seg}
\end{figure*}
\subsubsection{Retrieved $K$ Nearest Representations (RQ2)}
As the representations from the \encoder are high-dimensional vectors, we use t-Distributed Stochastic Neighbor Embedding (T-SNE) to perform dimensionality reduction and visualize them in a lower three-dimensional space. We randomly sampled 800 representations from the datastore and then selected the top 50 nearest representations for sensor 0 at timestep 13789. 
As illustrated in Fig. \ref{fig:retrival_embeding}, the retrieved similar representations are mapped to points that are close to each other and some of them are heavily overlapped, suggesting a higher degree of similarity and shared characteristics in representations. On the other hand, the points related to randomly sampled representations are distributed sparsely and far away from each other in the lower three-dimensional space. This indicates that the corresponding representations from these points are dissimilar to the sensor 0 at 13789. Additionally, Fig. \ref{fig:retrival_embeding} illustrates that the \encoder provides distinguishing representations from the MTS raw data and \Name has the ability to extract sparsely distributed and 
% useful, 
similar segments from the entire datastore. 
% The experiment setting is the same as section \ref{exp_set}.
\begin{figure}
 \centering
\includegraphics[width=0.6\linewidth]{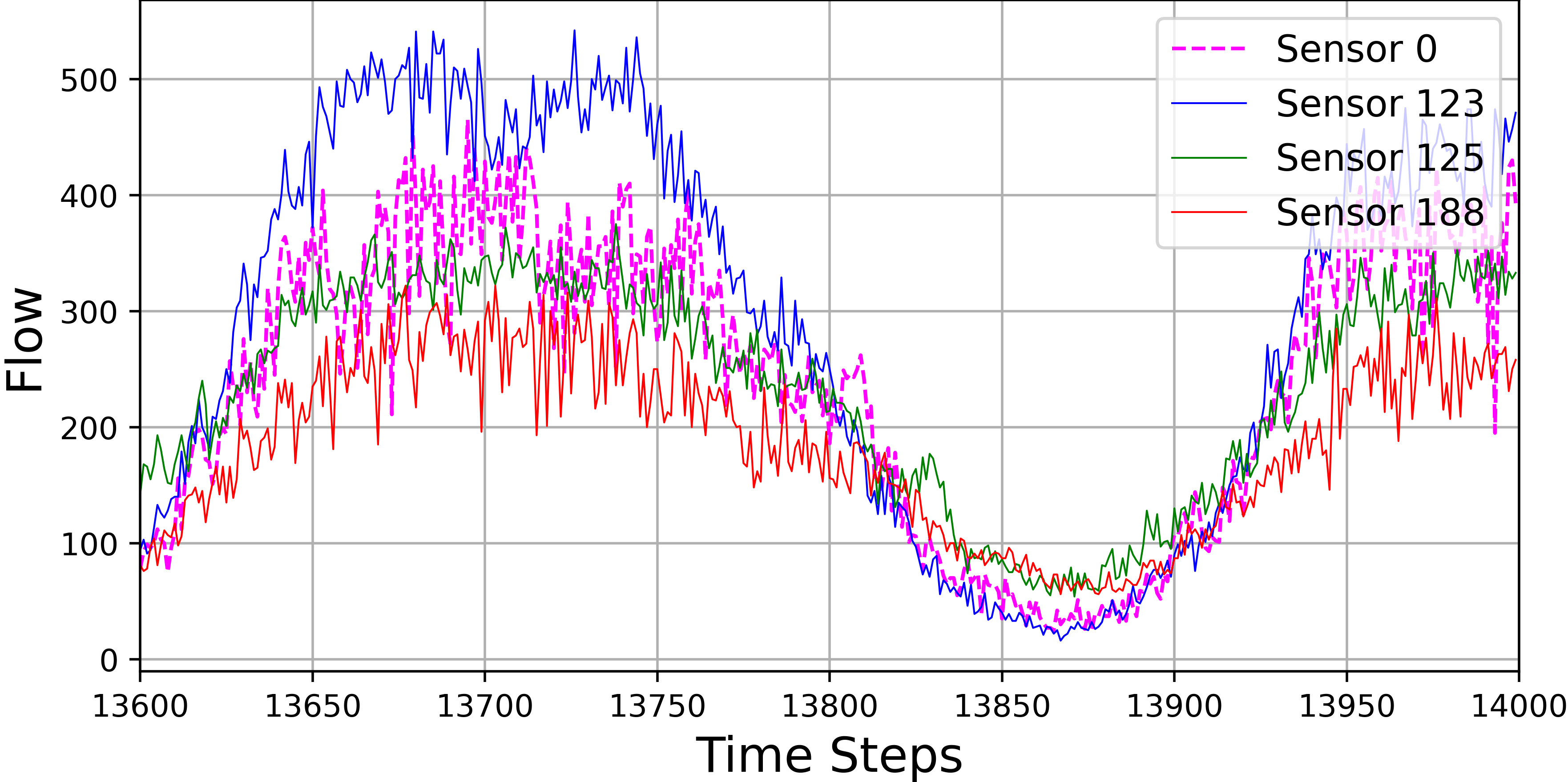}
 \caption{Retrieved corresponding sensors from \Name} 
    \label{vis_retr_node}
\end{figure}

\subsubsection{Retrieved $K$ Nearest Neighbours Segments and Series (RQ2)}
We then plot the retrieved segments (target future values) 
% and corresponding weight 
for the sensor 0 at timestep 13789. 
We first plot the top 5 historical segments from all retrieved segments in  Fig. \ref{vis_retr_seg}. Fig. \ref{vis_retr_seg} (a) shows the retrieved similar segments from sensor 0 at 7.7k, the time interval of the retrieved segments and target future is nearly 3 weeks (6050 timesteps), which indicates \Name can capture the cyclic patterns from the node itself over the long-term history. Fig. \ref{vis_retr_seg} (b) shows the retrieved similar segments from adjacent sensor 125 at 5.7k, the time interval of the retrieved segments and target future is 4 weeks (8064 timesteps) or with 45 minutes offset (8055 timesteps). These segments indicate \Name can find similar and cross-correlated segments from spatial domains and identify the adjacent sensors which could be useful in forecasting, and there could be a time lag for traffic flow to diffuse to its adjacent sensor. Fig. \ref{vis_retr_seg} (c)  illustrates that \Name finds a similar segment from the
non-adjacent sensor 123 at 3.7k to help the prediction, it also shows that cyclical patterns can emerge among sensors sharing similar characteristics in low dimensions, even if they are not physically located close to each other.
Moreover, all the potential retrieved segments have a similar trend and amplitude and would contribute to the target segments' forecasting.
Then we plot the corresponding sensors from the top 3 retrieved similar segments in Fig. \ref{vis_retr_node}. The retrieved series have similar long-term patterns, and there would exist inherent spatial-temporal dependency between these sensors. 
Apparently, \Name has a strong ability to identify similar segments  from MTS data and is interpretable because the data used to make the prediction can be directly inspected.

\begin{figure}
 \centering
 \includegraphics[width=\linewidth]{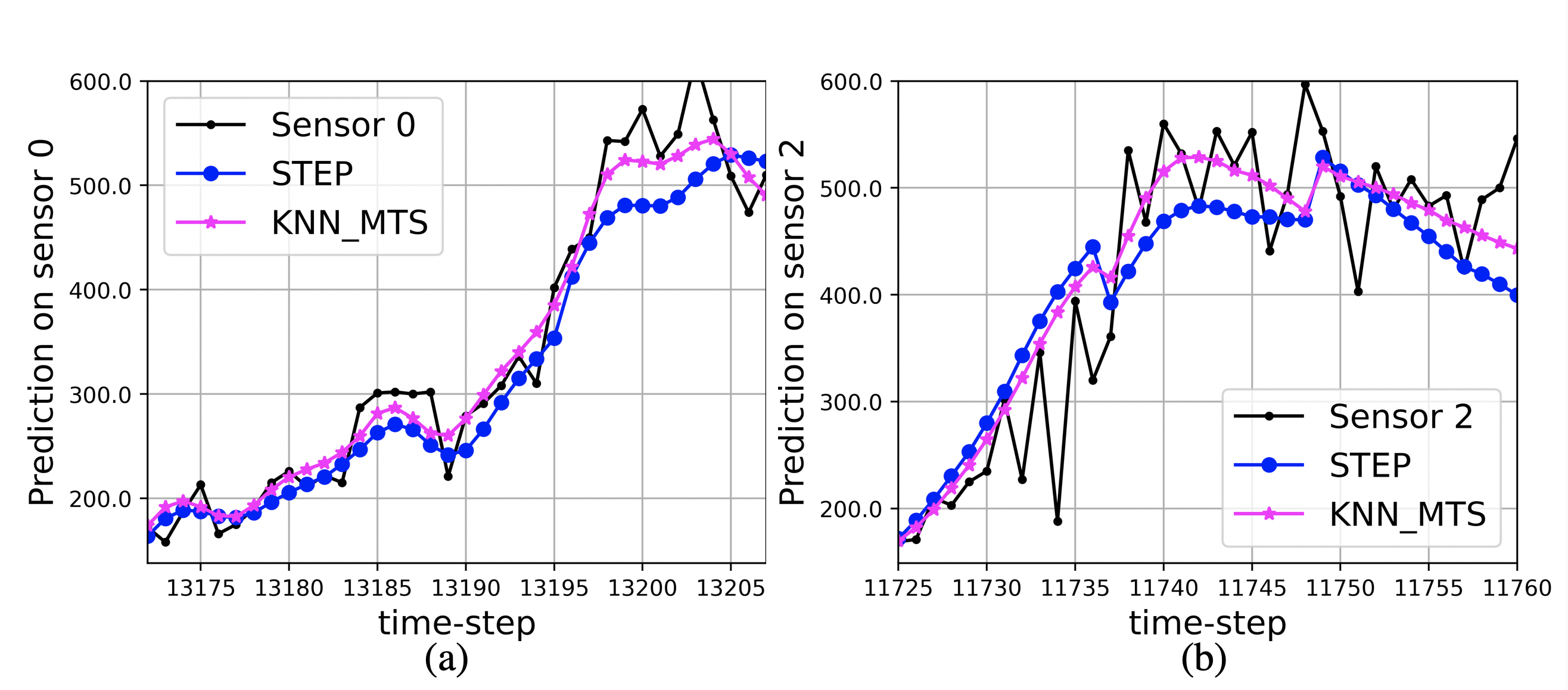}
    \setlength{\abovecaptionskip}{-0.2cm}
    \setlength{\belowcaptionskip}{-0.3cm}  
 \caption{Predictions from \Name and the state-of-the-art method STEP. (a) \Name outperforms the baseline by predicting rare patterns more accurately around the 13.2k time step. (b) \Name is more accurate in predicting rapid fluctuations around 11.7k.} 
\label{vis_compare0}
\end{figure}

\subsubsection{Visualization of Predictions (RQ3)}
We randomly selected a predicted time window from sensor 0 and sensor 2 to demonstrate the forecasting performance between \Name and on of the state-of-art method STEP.
% , and show the effectiveness of retrieval methods. 
As illustrated in Fig. \ref{vis_compare0}, the black curve represents the golden label for sensor 0 and 2, and the predictions of sensor using \Name and STEP are also depicted. We observe that \Name (purple) provides more accurate predictions and rapidly captures the wave of sensor 0 before the 13200-th time step than STEP (blue) in Fig. \ref{vis_compare0} (a).
% , especially in the time window around the 13200-th step for sensor 0. 
This is because the \Name framework 
% extends current MTS forecasting models by aggregating them with $k$-nearest neighbours and 
has the ability to extract sparse and useful information from the entire dataset, which is particularly helpful in remembering rare patterns from memory. In contrast, STEP can only do predictions based on fixed time scales. 
% Furthermore, \Name rapidly captures the wave of sensor 0 before the 13200-th time step, resulting in better predictions. 
Fig. \ref{vis_compare0} (b) also indicates that \Name is more accurate 
 % is closer to sensor 2's ground truth (black) than STEP (blue), 
 even when the real label fluctuates violently. 
 % We attribute this improvement to the assistance of the retrieved similar segments from the entire dataset. 
Based on these observations, \Name provides more robust and accurate predictions with the access to the entire dataset,
% multivariate long-term history data from the whole MTS dataset, \Name 
and can respond quickly to the sudden changes of each sensor and reduce the short-term effects from adjacent time series.
{
\begin{table*}[htbp!]
\renewcommand\arraystretch{1}
\centering
\caption{Ablation study on PEMS04 with different representations by changing the \encoder and choosing intermediate states. We retrieve k=50 neighbors and $\alpha=0.2$.}

\begin{tabular}{cccr|ccr|ccr}
\toprule
\hline

\multirow{2}*{\textbf{Variants}}    & \multicolumn{3}{c}{\textbf{Horizon 3}}     & \multicolumn{3}{c}{\textbf{Horizon 6}}   & \multicolumn{3}{c}{\textbf{Horizon 12}}\\ 
            \cmidrule(r){2-4} \cmidrule(r){5-7} \cmidrule(r){8-10}
        & MAE & RMSE & MAPE & MAE & RMSE & MAPE & MAE & RMSE & MAPE\\
\midrule

\Name & \textbf{17.08}$^*$  & \textbf{27.90}$^*$ & 
\textbf{11.52}\%$^*$ & \textbf{17.75}$^*$  & \textbf{29.32}$^*$ & \textbf{11.89}\%$^*$      & \textbf{18.68}  & \textbf{30.69} & \textbf{12.59}\%$^*$ \\
\encoder & 17.40  & 28.49  & 11.77\%  & 18.29  & 30.02  & 12.18\%  & 19.08 & 31.68  & 12.95\% \\
 \hline
\textit{\Name with GraphWaveNet}  & 17.61  & 28.79  & 11.77\%       & 18.46  & 30.20  & 12.16\%     &19.63  & 31.97  & 12.99\%\\ 
\textit{Graph WaveNet}   & 18.15  & 29.24  & 13.54\%        & 19.12  & 30.58  & 14.27\%   & 20.69  & 33.02  & 15.47\% \\
 \hline
\textit{\Name with Tsformer} & 17.34  & 28.18  & 11.73\%        & 18.17  & 30.08  & 12.09\%       & 19.03  & 31.27  & 12.86\% \\
 \textit{Tsformer}  & 17.72  & 28.97  & 11.83\%        & 18.37  & 30.26  & 12.21\%       & 19.11  & 31.77  & 12.98\% \\
  \hline
Linear Layer  Output    & 17.20  & 28.08  & 11.73\%       & 18.06  & 29.75  & 12.16\%     &19.03  & 31.05  & 12.99\% \\ 
Relu Layer  Output  &17.31 & 28.12 &11.90\%           &18.10 & 29.70 &12.19\% &18.99 &31.41 &12.98\% \\
\bottomrule
\end{tabular}
\label{tab:ablation}
\end{table*}
}

\section{Ablation Study}
\label{sec:abl}
While \Name is conceptually straightforward and requires no additional training, different types of representations could be used and  hyper-parameters are introduced for the nearest neighbour search. 
% We experiment with different choices here.
In this section, we conduct detailed experiments to verify the impacts of different components in \Name.
\subsection{Effect of Different Representations (RQ4)}

% In order to precisely find the most valuable time series from the whole MTS historical data, we explore the different ways of encoding spatial-temporal data to provide most powerful MTS representations. 

% We define three different types of MTS representations: \textbf{Univariate Temporal Representation, Multivariate Spatial-Temporal Representation}, and \textbf{Hybrid Representation}. In \Name, the hybrid version, which absorbs the first two representations 
% % univariate long-term and multivariate short-term history 
% shows the best forecasting performance. We name the hybrid representation generation method as Hybrid Spatial-Temporal Encoder (\encoder). 

% The current representations from STGNNs jointly capture the spatial and temporal patterns of MTS through graph neural networks and sequential models, which significantly improves the prediction accuracy. Due to the computation complexity of the models, most STGNNs only encode short-term historical MTS data, typically data from the past hour. 
In \Name, we use the representation from \encoder to build the datastore and retrieve for forecasting, which shows a significant improvement in the final performance by capturing both long-term temporal and short-term spatial-temporal dependencies from the entire dataset. Furthermore, \Name is also a general framework that can enhance almost any MTS forecasting model by absorbing the different MTS encoders. In this subsection, we inspect the generality of \Name framework by analyzing its performance using different types of representations, and we retrain the several base models to generate the corresponding dataset and forecasting based on the PEMS04.

To keep in line with our proposed \Name and to ensure a fair comparison, we select two representative base models which server as encoder in \Name: Graph Wavenet from STGNNs \cite{GWNet}  and Tsformer from transformer-based model using the segmentation \cite{shao2022pre}. Specifically, \textit{\Name with Graph WaveNet} uses the representation which focuses on the short-term spatial-temporal dependencies with $L=12$, and \textit{\Name with Tsformer} focuses on the long-term temporal dependencies with $L=2016$ and $P=168$. Additionally, we also present the performance of these base models for comparison.
% Since \Name is a general framework that can enhance almost any MTS forecasting model, we analyzed its performance by changing different representations. To do this, we dropped or replaced one component in \encoder as a new representation and retrained the model to generate the corresponding dataset based on the PEMS04. Specifically, \textit{\Name with Graph WaveNet} uses the representation from \encoder while cutting the $H_{long}$ long-term contextual information, while \textit{\Name with Tsformer} is the representation from \encoder without the $H_{short}$ short-term temporal and spatial representation. 
% For similarity search, 
% In \Name, we extract a representation using the final output of \encoder, which provides both long-term contextual information and short-term information with dependencies between time series. 
%%%%%%%%% steop here
As shown in Table \ref{tab:ablation}, \Name outperforms all its variants, demonstrating the superiority of using the representation from \encoder. Moreover, \Name with different base models significantly improve the accuracy of prediction across all horizons compared to direct predictions. \textit{\Name with Graph WaveNet} performs much better than original Graph WaveNet with similar segments from retrieval. Which shows the ability of \Name framework to search from the whole dataset compensates for the deficiency of a fixed-view in \textit{Graph WaveNet}, and some rare patterns with similar short-term representations could also be found to boost the final forecasting. What's more, direct prediction from \textit{Tsformer} is better than \textit{Graph WaveNet} since it contains long-term temporal dependencies information, which provides more accurate forecasting trends. 

% Apart from choosing representations generated from different backbone models, 
We also compare different choices from the intermediate states of \Name as representations without retraining the model. \Name utilizes representation $H_{hybrid}$ of the final state from the output layer (Relu layer).
% , and there is a Linear layer and another Relu layer before the final output layer.  
We compare the performance of two intermediate states (the second-to-last Linear layer and the third-to-last Relu layer) in the last two rows of Table \ref{tab:ablation}. While all the choices we tried were helpful, we achieved the largest improvement by using the model's final layer output. 
Repeating the experiment on the intermediate layer showed similar trends with slightly worse results, suggesting the final output might be focusing more on the prediction problem.
% , while the intermediate representation focused more on finding segments. 
Additionally, MTS forecasting is more complex due to the infinite space of possible values and multivariate dependency. Therefore, the final output layer gives a closer representation for prediction and more accurately retrieved target values.
% \ref{tab:ablation_key_transformer}.
\begin{comment}
Apart from choosing representations generated from different backbone models, 
% \Name also utilizes representation $H_{long}$ from the final state of the Transformer. 
we also compare several choices from the intermediate states of \Name to construct representations, with results shown in Table \ref{tab:ablation_key_transformer}. Transformer computes several intermediate states as shown in Equation \ref{eq:transformer}, and we compare the LayerNorm output of MHSA and FFN. While all the choices of $f$ we tried were helpful, we achieved the largest improvement by using the model's final layer output. Repeating the experiment on the intermediate transformer layer showed similar trends with slightly worse results, suggesting that the final output might be focusing more on the prediction problem, while the representation from the feedforward layer and the attention layer is more focused on finding segments. Additionally, MTS forecasting is more complex due to the infinite space of possible values and multivariate dependency. Therefore, the linear transformer of the final output layer gives a closer representation for prediction and more accurately retrieved target values than the non-linear layer from the MHSA and FFN output.
\end{comment}

% \input{tables/abla_key_normal}

\subsection{Training Efficiency of \Name (RQ5)}

\begin{figure}[htbp]
 \centering
 \includegraphics[width=1\linewidth]{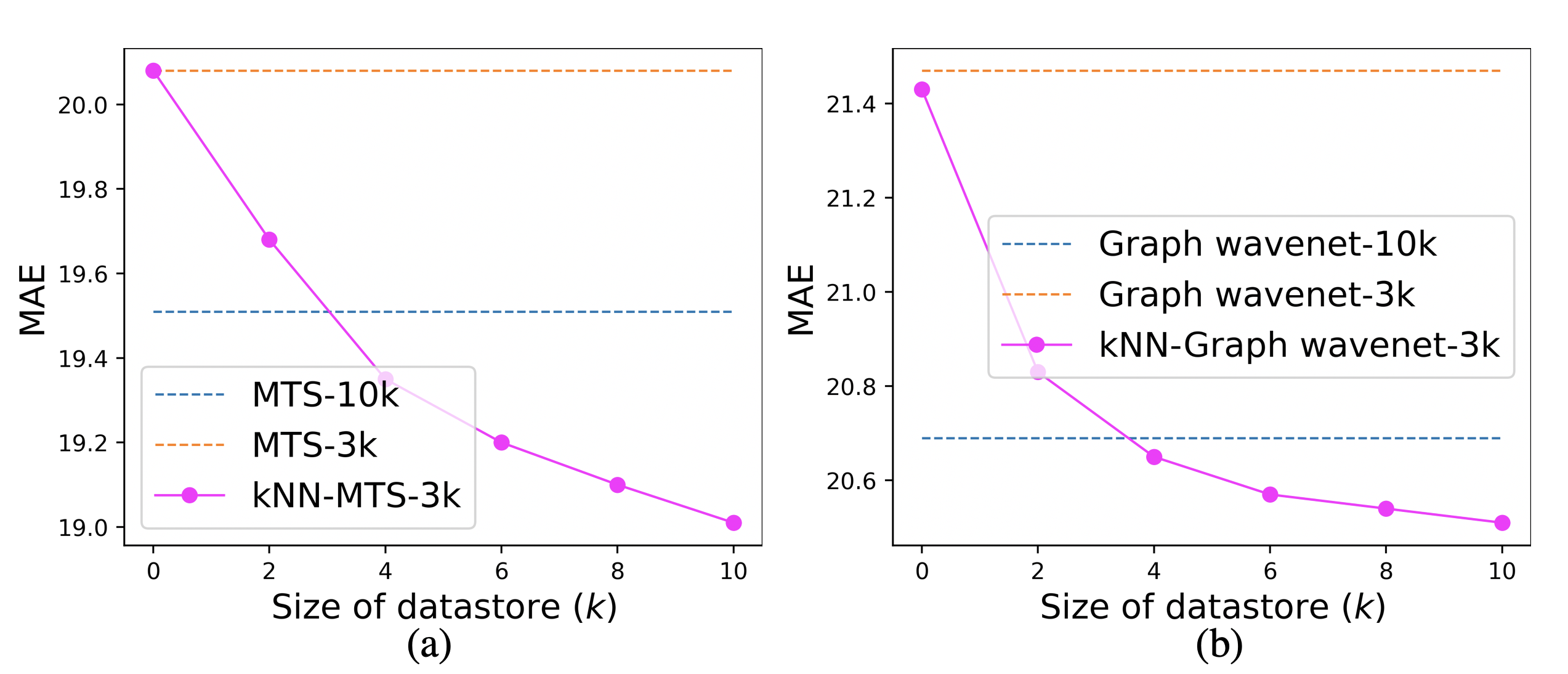}
 \caption{Varying the size of the datastore. (a) 
    % Increasing the datastore size monotonically improves performance, and has not saturated even at about 10k training data for each node. 
    \Name trained on 3k with a datastore of 2.5k already outperforms the MTS model trained on all 10k training data. (b) \Name with representation from Graph WaveNet trained on 3k with a datastore of 3.7k outperforms the Graph WaveNet model trained on all 10k training data.} 
 \label{exp_data_size}
\end{figure}

Provided with more training data, MTS modelling performance can be greatly enhanced, but at the cost of reduced training efficiency. This raises the question: can retrieving the nearest neighbours from the data be a substitute for training on it? Thus improving the performance without sacrificing training efficiency? To test this, we trained our \encoder using only 3k training data for each node and used it to build a datastore with increasing size, randomly sampled from the 10k training set. We then compared this approach, which we refer to as \Name-3k, to a vanilla \encoder trained on the entire 10k (MTS-10k) training set.
% We first experimented with creating a datastore from the same data used to train the \encoder. 
Fig. \ref{exp_data_size} (a) shows that \Name-3k improves prediction accuracy from 20.07 to a new state-of-the-art of 19.01 with increasing datastore on average MAE. It also has not saturated even at around 10k, suggesting that growing the datastore more could lead to further gains. Moreover, we use Graph WaveNet to generate the representations and build the datastore with different sizes, the experiments results can be found in  Fig. \ref{exp_data_size} (b). $k$NN-Graph WaveNet trained on 3k with a datastore of 3.7k also outperforms the Graph WaveNet  model trained on all 10k training data.
% , which shows that similar segments can improve prediction with arbitrary representations dramatically.
Furthermore, we found that training a model on 3k dataset for each node and using kNN search over a 10k dataset can outperform training the same model on all 10k dataset. This result suggests that, rather than training models on a larger dataset, we can use a smaller dataset to learn representations and augment them with \Name over larger quantities of data. This opens up a new path for efficiently using large dataset in MTS models and improving training efficiency. Furthermore, \Name is an expressive framework because it can use an arbitrary amount of data at test time to consistently improve performance. 

\subsection{Hyper-parameters Impacts (RQ6)}
\begin{figure}[htbp]
\centering
\includegraphics[width=\linewidth]{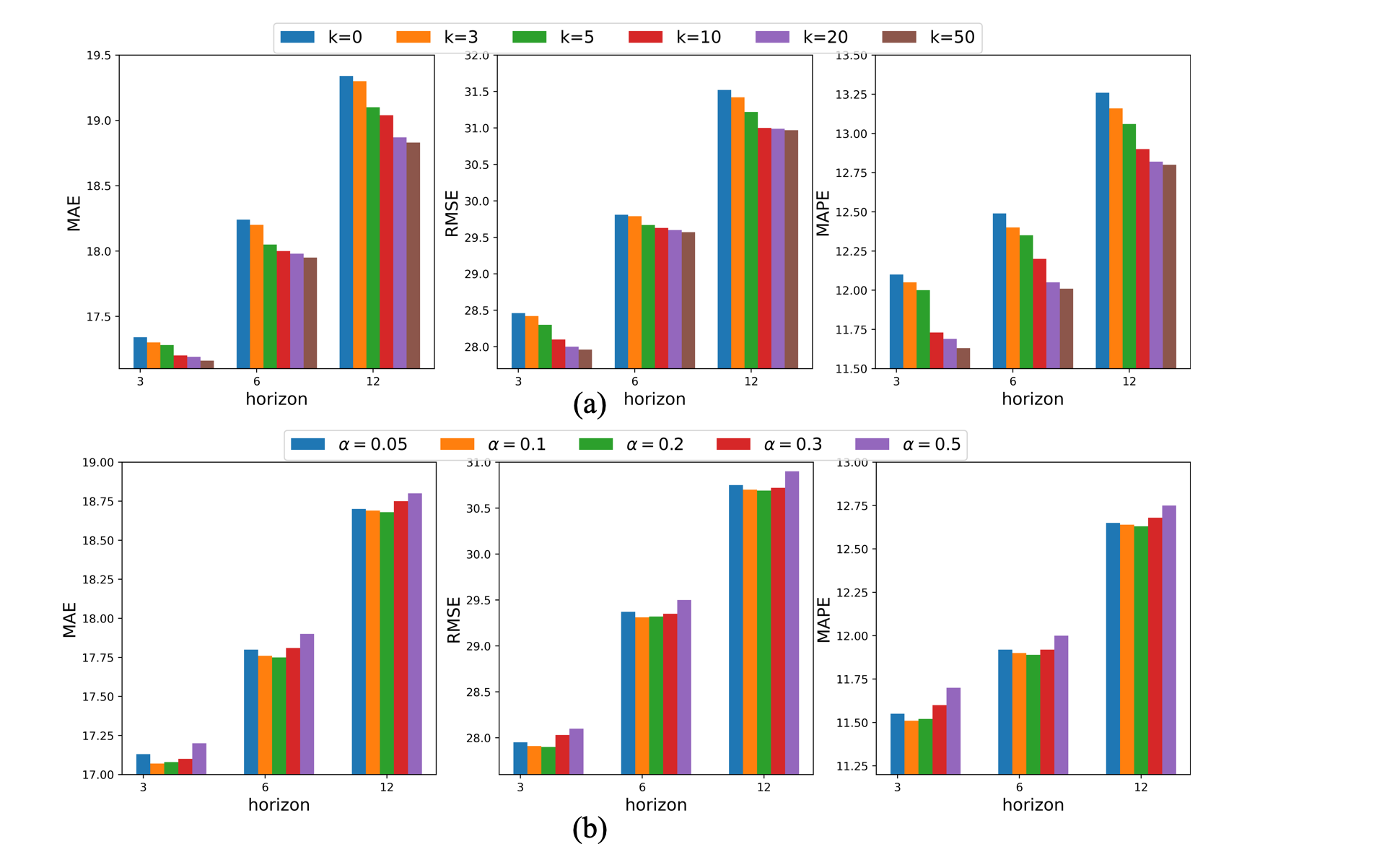}
\caption{Hyper-parameters Impacts. (a) Impact of $K$ retrieved similar historical segments. 
% Prediction performance monotonically improves as more similar segments are aggregated. 
(b) Effect of scaling parameter $\alpha$. }
% $\lambda = 0.5$ is optimal on PEMS04.}
\label{fig:kn ks hyper}
\end{figure}

We conduct experiments to analyze the impacts of two hyperparameters: the number of retrieved $K$ similar segments and scaling parameter $\alpha$.
In retrieval, each query returns the top-$K$ neighbours. Fig. \ref{fig:kn ks hyper} (a) shows that performance monotonically improves as more neighbours are returned, and suggests that even larger improvements may be possible with a higher value of $k$. Nonetheless, even a small number of neighbours $(K =10)$ is enough to achieve a new state of the art.
We use a parameter $\alpha$ to adjust the scaling of $\lambda$ which control the aggregation weight between the base model prediction $\hat{Y}^i$ and the retrieved similar targets values $\sum_{j\in K} w_{j}*Y_{j}$ with $K=50$. 
% , the use of $\lambda$ makes the predictions more robust in cases where there are no relevant cached examples. 
Fig.  \ref{fig:kn ks hyper} (b) shows that $\alpha=0.2$ is optimal on PEMS04 and the model is not sensitive to the values of $\alpha$ when it's in the range from 0.1 to 0.3.
% \vspace{-0.1cm}
\subsection{Computation and Space Cost (RQ7)}
\label{RQ7}
% In this subsection, we explore whether \Name can predict effectively and efficiently compared with the state-of-the-art baseline using their inference time. 
Although the \Name requires no further training given existing MTS forecasting models, it does add some other computational overheads. 
% The computation and space cost happens in building datastore and forecasting. 
When creating the datastore, storing the keys and values requires a single forward pass over the training set.
% , which amounts to a fraction of the cost of training for one epoch on the same examples. 
Taking PEMS04 dataset as an example, a single epoch of training over the entire dataset is 5 minutes on a signle GPU. In comparison, a single forward pass over the same dataset to save keys/values takes 3 minutes. Then, creating the datastore using \textsc{Faiss} takes 2 minutes. Hence, building the datastore is comparable to a single epoch of training. In addition, the saving of keys/values as well as creating the datastore are trivial to parallelize.
While the datastore component in \Name does require storage, it is not GPU based, which makes this storage very cheap. This is possible because the kNN component does not add any trainable parameters and requires no further parameter updates on the GPU, and it takes about 1 GB of disk space for the PEMS04 datastore.
 % building the cache with 10k samples for each node takes roughly 380 seconds on a single CPU. While the cost of building a large cache grows linearly in the number of training data, it is trivial to parallelize and requires no GPU-based training. 
 % \textcolor{blue}{For the space cost, }. The main expense of constructing the datastore is a single forward pass over all examples in the datastore, which is only a fraction of the cost of training on the same examples for one epoch.  During forecasting, retrieving $K$ keys from a datastore containing tens of millions of items leads to a generation speed that is two orders of magnitude slower than the base MTS models.  In practice, the search and storing of key-value pairs in datastore \textcolor{blue}{add space cost, which is smaller than a big model.}
In forecasting, the average inference speed of \Name required for each epoch on the whole validation set (307 sensors and retrieving 50 segments for each) is roughly 113 points per second per sensor on one GPU. This speed is easily fast enough for most applications, albeit slower than STEP, which can be roughly 128 points. The inference speed is 240 points per second per sensor when we choose Graph WaveNet as the base model called $k$NN-Graph WaveNet,  which achieves at least 5\% better performance compared with vanilla Graph WaveNet with a speed of 285 points per second per sensor. Moreover, developing faster nearest-neighbor search tools remains an active area of research \cite{guo2020accelerating} and can improve the inference speed.

\vspace{-0.2cm}
\section{Related Work}
\label{sec:related}
We briefly review the related work from two aspects: the MTS forecasting and STGNNs.

\vspace{-0.1cm}
\subsection{MTS Forecasting}
The field of time series forecasting has been studied for many decades. One of the famous and widely used traditional methods for time series forecasting is the auto-regressive integrated moving average (ARIMA) model. This popularity arises from its strong statistical properties and its adaptability in integrating various linear models, including auto-regression (AR), moving average (MA), and auto-regressive moving average components. However, ARIMA encounters limitations due to its high computational complexity, making it impractical for modeling multivariate time series (MTS) data.
Then, extensions of ARIMA have been developed, such as the  vector auto-regression moving average (VARMA) models. Another approach involves using Gaussian processes (GP) \cite{xie2010gaussian_traffic}, a Bayesian method designed to model distributions across a continuous domain of functions. GPs serve as a valuable prior within Bayesian inference and have found application in time series forecasting. Nonetheless, these approaches often rely on the strict assumption of stationarity and may struggle to capture non-linear dependencies among variables.% DCRNN

In contrast, recent deep learning-based methods are free from these limitations and demonstrate better performances. Most of existing works rely on LSTM and GRU to capture temporal dependencies \cite{MTS_survey,he2024distributionalshift_INNLS}.  
% Compared with those RNN-based approaches, dilated 1D convolutions \cite{wu2020connecting} are able to handle long-range sequences. but may cause the loss of local information, which brings in negative effects on modeling short-term dependencies. 
Some other efforts exploit TCNs and self-attention mechanism \cite{self-attention-ts} to model long time series efficiently. Models like LSTNet \cite{LSTnet} and TPA-LSTM \cite{tpa-lstm} have been developed to capture discrete temporal dynamics and local spatial correlations between time series using a combination via RNNs and CNNs. To address the parallelization issue in RNNs, methods built on Transformer \cite{zhou2021informer, patchTst, zhang2022crossformer} demonstrate a better efficiency and forecasting ability. For example, Informer \cite{zhou2021informer} achieves similar results by introducing a variant of the vanilla Transformer. 
% More recently, PatchTST \cite{patchTst} utilizes channel-independent and subseries-level patches to capture local semantic information and attend longer history. 
To mine and distinguish patterns from the long-term history, \cite{JointSpatiotemporal_INNLS} uses domain knowledge to design a fuzzy cognitive map to extract the useful features with different spatial resolutions and the spatiotemporal dynamics from long-term history
% , at the cost of quantities of expert knowledge in 
fuzzy cognitive map design. 
\cite{STWave_TKDE} utilizes the historical trend knowledge as a self-supervised signal to teach overall temporal information through a contrastive loss. PatchTST \cite{patchTst} and Crossformer \cite{zhang2022crossformer} 
attempt to convert the long-term history into segment embeddings through cross-time-level and cross-dimension-level Transformer. 

% However, all those methods have not explicitly modeled the pairwise dependencies between variables,limiting their effectiveness in forecasting multi variate time series.
% stgnn
% attention to get better diffusion
% step autoformer short term- longterm work
% graph learning
% \subsection{Spatial-Temporal Graph Neural Networks}
% The successful development of deep learning (DL), which is based on artificial neural networks, has revolutionized time series forecasting tasks~\cite{Innovation}. 
\vspace{-0.2cm}
\subsection{Spatial-temporal Graph Neural Networks}
The GNN~\cite{2017GCN, 2016ChebNet} has been introduced as a new deep learning paradigm for learning non-Euclidean data by applying graph analysis methods. 
% Specifically, 
STGNNs is based on the concept of simultaneously modeling spatial and temporal dependencies to deal with graph problems. 
% The accuracy of multivariate time series forecasting has been largely improved by artificial intelligence~\cite{Innovation}, especially deep learning techniques.
% Among these techniques, Spatial-Temporal Graph Neural Networks~(STGNNs) are the most promising methods, 
% which combine Graph Neural Networks (GNNs)~\cite{2017GCN, 2016ChebNet} and sequential models~\cite{2014GRU, 2014Seq2Seq} to model the spatial and temporal dependency jointly.
There are two trends for STGNN methods. The first is the graph convolutional recurrent neural network. 
DCRNN~\cite{2017DCRNN}, STMetaNet~\cite{2019STMetaNet}, AGCRN~\cite{2020AdaptiveGCRN}, and TGCN~\cite{2019TGCN} combine diffusion convolution networks and recurrent neural networks~\cite{2014GRU, 2014Seq2Seq} with their variants. 
They follow the seq2seq framework~\cite{2014Seq2Seq}  and adopt an encoder-decoder architecture to forecast step by step. In particular, the main limitations of these RNN-based networks are computational and memory bottlenecks, which make it difficult to scale up with large networks.
% .
The second trend is the fully spatial-temporal graph convolutional network which aims to address the computational problem caused by using recurrent units. Graph WaveNet~\cite{GWNet}, MTGNN~\cite{2020MTGNN}, STGCN~\cite{stgcnyu2017spatio}, and StemGNN~\cite{2020StemGNN} combine graph convolutional networks and gated temporal convolutional networks with their variants. 
These methods are based on convolution operation, which facilitates parallel computation.
Moreover, the attention mechanism is widely used in many methods for modeling spatial temporal correlations, such as GMAN~\cite{2020GMAN}, TraverseNet\cite{TraverseNet_INNLS} and ASTGCN~\cite{2019ASTGCN}. 
MAGNN \cite{chen2023multi_scale} designs an adaptive graph learning module to explore the
multivariable dependencies, and more recent efforts are focused on utilizing the mutil-time scales features from MTS \cite{Multiscale_INNLS, MR-Transformer_INNLS, my2024dmg_infor_sicence, STWave_TKDE} in modelling.
Although STGNNs have made significant progress, the complexity of the above STGNNs is still high because they need to deal with both temporal and spatial dependency at every time step. Hence, STGNNs are limited to taking short-term multivariate historical time series as input, such as the classic settings of 12 time steps. Recently, STEP and  PatchTST~\cite{shao2022pre,patchTst} find that long-term historical data of each time series also helps the final forecasting.  Unfortunately,
current methods only focus on a finite length of input data and cannot identify useful patterns from
the entire dataset, or struggle with data having sparsely and discontinuously distributed correlations among variables.
% , and they also struggle with data that have sparsely and discontinuously distributed correlations among variables.
% However, STEP still ignores the potentially valuable patterns underlying multiple long-term historical data. Different from STEP the proposed \Name in this paper mainly focuses on mining crucial representations from multivariate long-term history other than a single time series. 

\begin{comment}
Besides, most STGNNs are developed based on defined graph structures for the propagation process.
% , which cannot be directly extended to multivariate problems. 
In this regard, recent studies have introduced graph learning layers to automatically learn hidden spatial dependencies among nodes by data-driven perspectives~\cite{2018NRI, 2019LDS, 2021GTS}.  LDS~\cite{2019LDS} models the edges as random variables whose parameters are treated as hyperparameters in a bilevel learning framework. The random variables parameterize the element-wise Bernoulli distribution from which the adjacency matrix $\mathbf{A}$ is sampled. GTS~\cite{2021GTS} introduces a neighborhood graph as a regularization that improves graph quality and reformulates the problem as a uni-level optimization problem. The adaptive graph learned captures underlying dependency among neighbors and improves the final forecasting performance.
\end{comment}

\vspace{-0.2cm}
\section{Conclusion}
\label{sec:conclusion}
In this paper, we introduce a simple yet effective framework called \Name, which extracts sparse distributed but similar patterns using nearest neighbor retrieval based on representations from the entire dataset. 
\Name can be applied to any pretrained MTS model without further training and is a highly expressive and interpretative model that consistently improves performance.
We also propose a \encoder for \Name which can capture both long-term temporal and short-term spatial-temporal dependencies, and is shown to provide accurate representation for \Name for better forecasting. Extensive experiments on several real-world datasets show a significant improvement in the forecasting performance of \Name.   
Future work should further improve the efficiency and generalizability of this framework, including but not limited to down-sampling frequent target time series in the datastore, considering the distribution shifts of new samples, or enhancing representations and retrieval accuracy through contrastive learning, showing better application prospects and opening up a new path
for efficiently using the large dataset in MTS models.

% \section*{Acknowledgment}

\ifCLASSOPTIONcaptionsoff
  \newpage
\fi



%\begin{thebibliography}{1}


\bibliographystyle{IEEEtran}
\bibliography{ref}
% \bibliography{IEEEabrv,Bibliography}
%\end{thebibliography}
% \newpage
\begin{IEEEbiography}[{\includegraphics[width=1in,height=1.25in,clip,keepaspectratio]{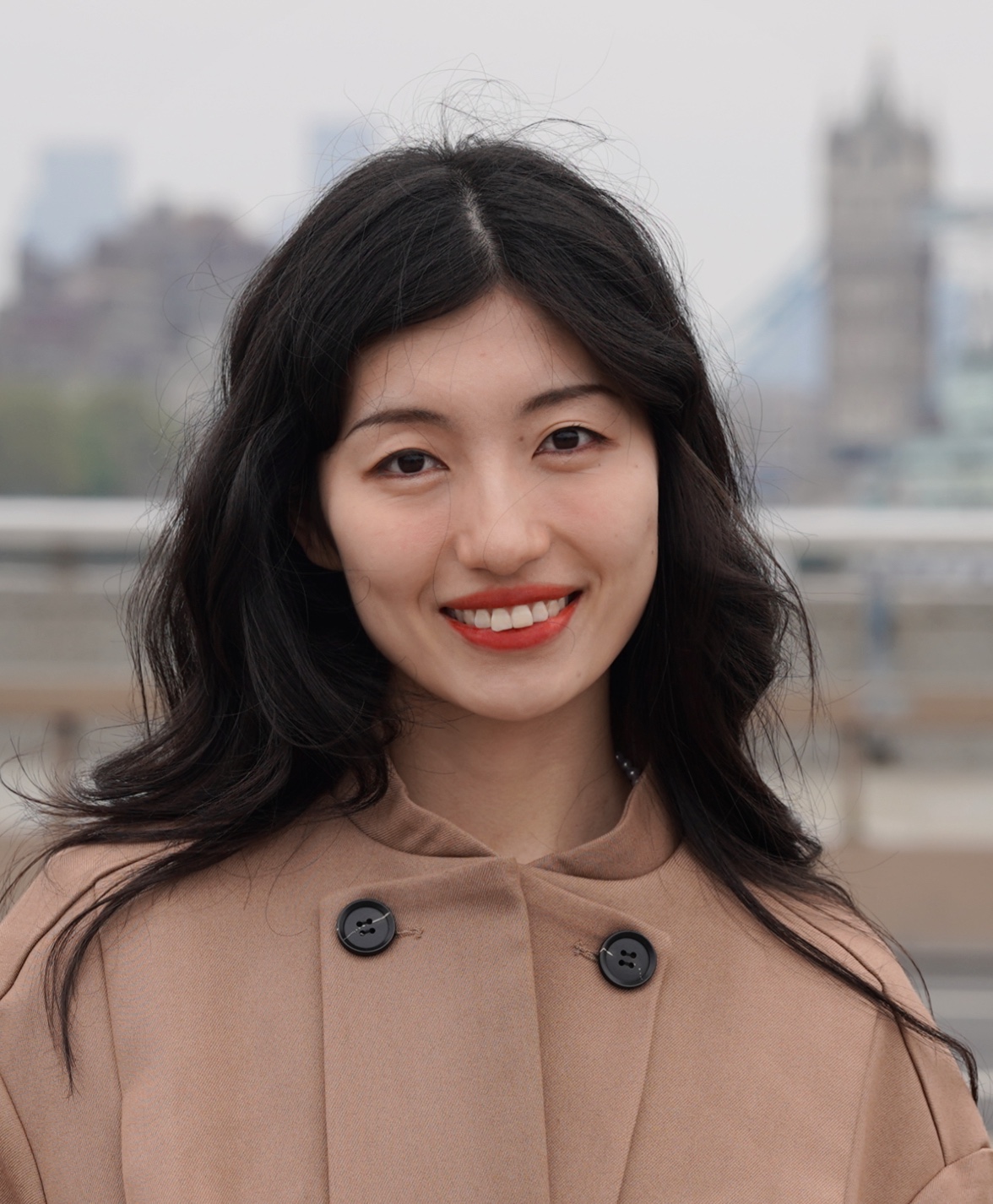}}]{Huiliang Zhang} is currently a Ph.D. candidate in the Department of Electrical and Computer Engineering at McGill University. She received her M.E. degree from Peking University, Beijing, China, in 2020 and her B.E. degree from Xidian University, Xi'an, China, in 2017.
She is interested in time series analysis, spatiotemporal data and graph modelling, and other machine learning and statistics solutions for research and applications in smart grids and intelligent transportation systems.
\end{IEEEbiography}
\begin{IEEEbiography}[{\includegraphics[width=1in,height=1.25in,clip,keepaspectratio]{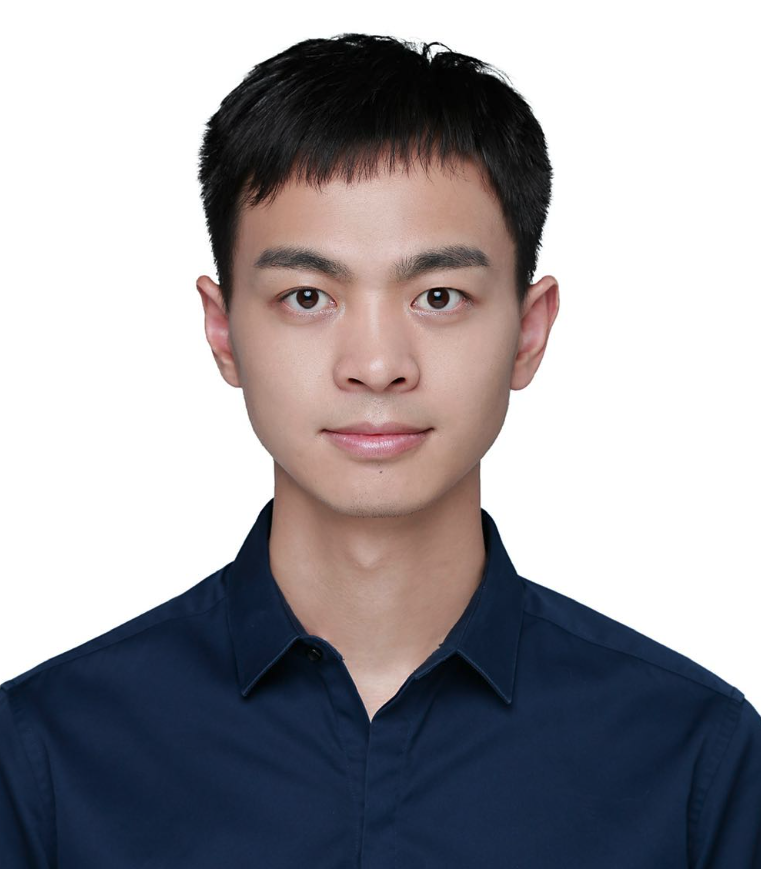}}]{Ping Nie} is an Applied Scientist. He received his Master's Degree from Peking University in 2020. His research interests include Recommendation Systems, Information Retrieval, Natural Language Processing, and Time Series Forecasting. He has published several publications on top Conferences and Journals such as ACL, SIGIR, NeurIPS, EMNLP, CIKM and TOIS. 
\end{IEEEbiography}
% \vspace{-1in}
% \newpage
\begin{IEEEbiography}[{\includegraphics[width=1in,height=1.25in,clip,keepaspectratio]{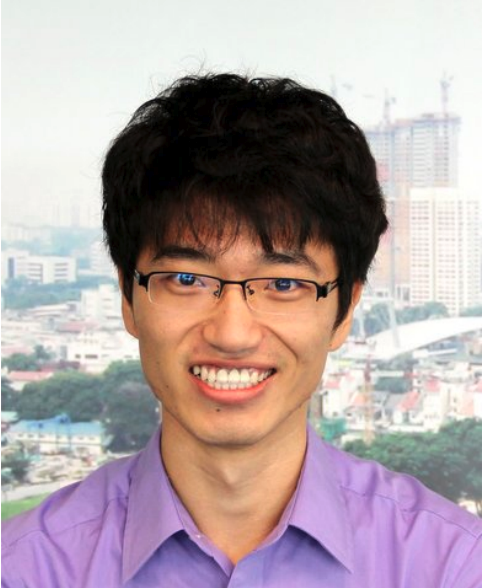}}]{Lijun Sun}
received B.S. degree in Civil Engineering
from Tsinghua University, Beijing, China, in 2011
and Ph.D. degree in Civil Engineering (Transportation) from the National University of Singapore in
2015. He is currently an Associate Professor with the
Department of Civil Engineering at McGill University, Montreal, QC, Canada. His research centers on
intelligent transportation systems, machine learning,
spatiotemporal modeling, travel behavior, and agent-based simulation. He is an Associate Editor of Transportation Research Part C: Emerging Technologies.
\end{IEEEbiography}
% \vspace{-1in}
\begin{IEEEbiography}[{\includegraphics[width=1in,height=1.25in,clip,keepaspectratio]{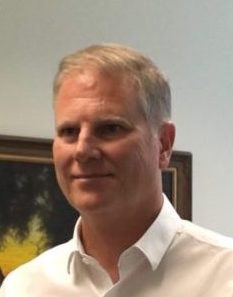}}]{Benoit Boulet} (S’88–M’92–SM’07), P.Eng., Ph.D., is Professor in the Department of Electrical and Computer Engineering at McGill University which he joined in 1998, and Director of the McGill Engine, a Technological Innovation and Entrepreneurship Centre. He is Associate Vice-Principal of McGill Innovation and Partnerships and was Associate Dean (Research \& Innovation) of the Faculty of Engineering from 2014 to 2020. Professor Boulet obtained a Bachelor's degree in applied sciences from Universit\'{e} Laval in 1990, a Master of Engineering degree from McGill University in 1992, and a Ph.D. degree from the University of Toronto in 1996, all in electrical engineering. He is a former Director and current member of the McGill Centre for Intelligent Machines where he heads the Intelligent Automation Laboratory. His research areas include the design and data-driven control of electric vehicles and renewable energy systems, machine learning applied to biomedical systems, and robust industrial control.
\end{IEEEbiography}
%\begin{biography}[{\includegraphics[width=1in,height=1.25in,clip,keepaspectratio]{mshell}}]{Michael Shell}

% ==== SWITCH OFF the BIO for submission
% ==== SWITCH OFF the BIO for submission

\end{document}